\documentclass[english]{article}
\usepackage[T1]{fontenc}
\usepackage[latin9]{inputenc}
\usepackage{geometry}
\usepackage[unicode=true,
 bookmarks=false,
 breaklinks=false,
 pdfborder={0 0 1},
 colorlinks=false]{hyperref}
\hypersetup{colorlinks,
linkcolor=red,
anchorcolor=blue,
citecolor=blue,
urlcolor=blue}
\usepackage{bm}
\usepackage{amsmath}
\usepackage{mathtools}
\usepackage{amssymb}
\usepackage{amsthm}
\usepackage{comment}
\usepackage{natbib}
\usepackage{booktabs}
\usepackage{graphicx}
\usepackage{algorithm}
\usepackage{algpseudocode}
\usepackage{float}
\usepackage{multirow}
\usepackage{dsfont}
\usepackage{tcolorbox}
\usepackage{tabulary}
\usepackage{colortbl}
\usepackage{color}
\usepackage[outdir=./]{epstopdf}
\usepackage{enumitem}

\definecolor{gen}{RGB}{0,0,200}
\definecolor{cc}{RGB}{231,117,0}

\allowdisplaybreaks

\newcommand{\ind}{\mathds{1}}
\newcommand{\defn}{\coloneqq}

\newcommand{\cS}{\mathcal{S}}

\newcommand{\cD}{\mathcal{D}}

\newcommand{\cF}{\mathcal{F}}

\newcommand{\cP}{\mathcal{P}}

\newcommand{\cM}{\mathcal{M}}
\newcommand{\cW}{\mathcal{W}}
\newcommand{\cL}{\mathcal{L}}

\newcommand{\bE}{\mathbb{E}}
\newcommand{\bP}{\mathbb{P}}
\newcommand{\bR}{\mathbb{R}}

\newcommand{\bX}{\mathbb{X}}

\newcommand{\diff}{\,\mathrm{d}}

\newcommand{\numpf}[2]{\overset{(\mathrm{#1})}{#2}}
\newcommand{\ol}{\overline}
\newcommand{\wh}{\widehat}
\newcommand{\wt}{\widetilde}

\newcommand{\data}{\mathsf{data}}

\newcommand{\veps}{\varepsilon}

\newcommand{\setc}{\mathrm{c}}

\newcommand{\mask}{\mathsf{M}}

\newcommand{\rel}{\mathsf{rel}}

\DeclareMathOperator*{\argmax}{arg\,max}

\newcommand{\KL}{\mathsf{KL}}

\newcommand{\ptI}{\mathtt{I}}
\newcommand{\ptH}{\mathtt{H}}

\title{Confidence-Based Decoding is Provably Efficient \\ for Diffusion Language Models under Random Ordering\footnotetext{Corresponding author: Gen Li.}}

\author{Changxiao Cai\thanks{Department of Industrial and Operations Engineering, University of Michigan, Ann Arbor, USA; Email: \href{mailto:cxcai@umich.edu}{cxcai@umich.edu}.}
\and
Gen Li\thanks{Department of Statistics and Data Science, The Chinese University of Hong Kong, Hong Kong; Email: \href{genli@cuhk.edu.hk}{genli@cuhk.edu.hk}.
}}
\date{\today}

\begin{document}

\theoremstyle{plain} 
\newtheorem{lemma}{\bf Lemma} 
\newtheorem{proposition}{\bf Proposition}
\newtheorem{theorem}{\bf Theorem}
\newtheorem{corollary}{\bf Corollary} 
\newtheorem{claim}{\bf Claim}

\theoremstyle{remark}
\newtheorem{assumption}{\bf Assumption} 
\newtheorem{definition}{\bf Definition} 
\newtheorem{condition}{\bf Condition}
\newtheorem{property}{\bf Property} 
\newtheorem{example}{\bf Example}
\newtheorem{fact}{\bf Fact}
\newtheorem{remark}{\bf Remark}

\maketitle 

\begin{abstract}
    Diffusion language models (DLMs) have emerged as a promising alternative to autoregressive (AR) models for language modeling, allowing flexible generation order and parallel generation of multiple tokens. However, this flexibility introduces a challenge absent in AR models: the \emph{decoding strategy}---which determines the order and number of tokens generated at each iteration---critically affects sampling efficiency. Among decoding strategies explored in practice, confidence-based methods, which adaptively select which and how many tokens to unmask based on prediction confidence, have shown strong empirical performance. Despite this success, our theoretical understanding of confidence-based decoding remains limited.
    In this work, we develop the first theoretical analysis framework for confidence-based decoding in DLMs. We focus on an entropy sum-based strategy that continues unmasking tokens in a uniformly random order within each iteration until the cumulative entropy exceeds a threshold, and show that it achieves $\varepsilon$-accurate sampling in KL divergence with an expected number of iterations $\widetilde O(H(X_0)/\varepsilon)$, where $H(X_0)$ denotes the entropy of the target data distribution and $\widetilde O(\cdot)$ denotes an asymptotic upper bound up to constant and logarithmic factors. Notably, this strategy yields substantial sampling acceleration when the data distribution has low entropy relative to the sequence length, while automatically adapting to the intrinsic complexity of data without requiring prior knowledge or hyperparameter tuning. Overall, our results provide a theoretical foundation for confidence-based decoding and may inform the design of more efficient decoding strategies for DLMs.
\end{abstract}

\medskip


\tableofcontents

\section{Introduction}
\label{sec:intro}
Diffusion language models (DLMs) \citep{sahoo2024simple,shi2024simplified} have recently emerged as a compelling alternative to the autoregressive (AR) paradigm for language modeling.
Unlike AR models, DLMs use a non-causal generation mechanism that can produce multiple tokens in parallel at each iteration and relax the rigid left-to-right constraint. 
The parallel generation capability offers strong potential for sampling acceleration, while the non-causal structure also makes DLMs well suited to tasks that benefit from bidirectional reasoning or the enforcement of global constraints  \citep{nie2025large,prabhudesai2025diffusion,li2026convergeflow}.
At scale, DLMs have been shown to achieve performance competitive with AR models \citep{ye2025dream,song2025seed,you2025llada,zhu2025llada,bie2025llada2,labs2025mercury}.
Beyond their role as standalone generators, DLMs have also been used as speculative drafters to accelerate AR inference through parallel token generation \citep{chen2026dflash,cheng2026dspark,zhao2026training}.

State-of-the-art DLMs are built upon \textit{masked diffusion models} (MDMs) \citep{austin2021structured,campbell2022continuous,lou2023discrete}, a class of discrete diffusion models that augment the vocabulary with an additional absorbing state, referred to as \textit{mask}.
An MDM is defined through two complementary processes. In the forward (masking) process, tokens in a sequence drawn from the target data distribution are progressively replaced by the mask until the sequence becomes fully masked.
The reverse (unmasking) process then seeks to invert this corruption by iteratively recovering masked tokens, thereby transforming a fully masked sequence into a fresh sample from the data distribution.

Central to the reverse process is a \textit{mask predictor}, which learns the conditional distributions of masked tokens given the currently unmasked context at various masking levels \citep{lou2023discrete,ou2024your,zheng2024masked}.
The reverse dynamics are implemented by repeatedly sampling token values from these conditionals to impute masked positions, thereby moving from a fully masked sequence toward a data-like sample.
In this sense, the mask predictor in MDMs serves as a discrete analog of the score estimator in continuous diffusion models \citep{song2020score}.
As the joint distribution of multiple masked tokens is generally intractable in high-dimensional settings, practical mask predictors are typically trained to learn the conditional marginals of each masked token, and use the product of per-token conditional marginals as an approximation \citep{nie2025large,wu2025fast}.
In particular, a neural network-based mask predictor can output, in one forward pass, conditional marginal distributions for all currently masked positions.

While this factorized approximation enables parallel generation of DLMs, it also introduces an inevitable bias in the sampling stage---simultaneously unmasking multiple tokens in one iteration ignores their statistical dependence conditioned on the currently revealed context.
At one extreme, unmasking all tokens in a single iteration incurs an error comparable to the discrepancy between the target distribution and the product of its marginals. At the other extreme, generating tokens one-by-one recovers the standard AR approach, which removes the factorization bias but sacrifices the parallel generation advantage of DLMs.

Therefore, the sampling speed and generation quality of DLMs depend critically on the decoding (or unmasking) strategy used at the inference stage, which determines both which positions to unmask and how many tokens to reveal at each iteration.

\paragraph{Parallel decoding strategies.}

Decoding strategies used in practice can be broadly categorized into two families.
\begin{itemize}
    \item \textbf{Uniform decoding strategies.} This class of decoding strategies prescribes a fixed total iteration count $T$ and a step-size schedule $s_1,s_2\dots,s_T$ with $s_t \geq 1$ and $\sum_{t=1}^T s_t = L$ (e.g., 
    cosine schedules \citep{shi2024simplified} or log-linear schedules \citep{sahoo2024simple,lou2023discrete}).
    At each iteration $t$, one samples a subset of size~$s_t$ uniformly at random from the remaining masked positions and then unmasks the tokens in the sampled subset.
    Note that this family of schemes treats all masked positions uniformly and thus belongs to the broader class of non-adaptive decoding strategies.

     \item \textbf{Confidence-based decoding strategies.} This collection of decoding strategies leverages the fact that a mask predictor provides not only a distribution but also a natural confidence signal for each token prediction. 
    It is thus natural to use prediction confidence to guide unmasking decisions at each iteration, with newly revealed tokens providing additional context in later iterations
     \citep{chang2022maskgit,zheng2023reparameterized,gong2024scaling}. In practice, confidence is often quantified by the entropy of the predicted distribution \citep{ben2025accelerated}, the mass of the most likely token \citep{nie2025large,yu2025dimple,wu2025fast}, or the likelihood gap between the two most likely tokens \citep{kim2025train}, leading to a range of effective heuristics. Compared to uniform decoding strategies, these schemes determine the decoding order and set size in an adaptive manner.
\end{itemize}

In contrast to the rapid development of empirical decoding strategies, our theoretical understanding of their sampling efficiency remains limited.
For uniform random schedules, substantial recent progress has been made in establishing theoretical guarantees for sampling performance \citep{li2025breaking,chen2025optimal,lavenant2025error,dmitriev2026efficient,zhao2026adaptation}.
We defer a detailed discussion to Section~\ref{sec:related-work}.
By comparison, despite their strong empirical performance, confidence-based decoding remains far less understood theoretically.
The fundamental difficulty lies in their inherently adaptive decoding dynamics: the set of tokens unmasked at each iteration depends on the \textit{realization} of the tokens generated so far, through the corresponding confidence levels produced by the mask predictor.
This adaptivity raises several fundamental but unanswered questions. For instance, if the total number of iterations is not predetermined, how many iterations are needed to unmask all tokens? What is the resulting sampling error? Can these schemes provably adapt to the (unknown) intrinsic complexity or dependence structure of the data distribution?

Such a theory-practice gap motivates the central question of our study:
\begin{center}
\textit{Are confidence-based decoding strategies in DLMs provably efficient?}
\end{center}

\subsection{Our contributions}

In this work, we take an initial step toward answering this question by developing sampling convergence guarantees for confidence-based decoding in DLMs, with the goal of demystifying their practical effectiveness.

We focus on a canonical confidence measure---entropy---and investigate entropy-based decoding that selects unmasking positions and set sizes based on the entropy of the predicted distribution at each masked position.
Concretely, we study a representative \textit{entropy sum-based strategy} that, at each iteration, unmasks tokens from the remaining masked positions in a uniformly random order until the running entropy sum of the newly unmasked tokens exceeds a prescribed threshold, thereby revealing as many tokens as possible per iteration while keeping the cumulative prediction uncertainty controlled.

We demonstrate that $\veps$-accuracy in  Kullback-Leibler (KL) divergence can be attained with an expected number of iterations that scales as
\begin{align*}
    \wt O\bigg(\frac{H(X_0)}{\veps}\bigg),
\end{align*}
where $H(X_0)$ is the entropy of the data distribution and $\wt O(\cdot)$ hides a logarithmic factor in the sequence length $L$.
This result implies a sublinear iteration complexity in the sequence length $L$ when the data distribution has low entropy $H(X_0) \ll L$, thereby achieving provable sampling speedup compared to the $\Theta(L)$ iteration complexity of the AR decoding or naive uniform parallel decoding. Here, $\Theta(\cdot)$ denotes matching asymptotic upper and lower bounds up to constant factors.
Notably, this scheme does not require any prior knowledge of the data distribution such as its entropy, and thus automatically adapts to the intrinsic low-complexity data structure to achieve sampling acceleration.

To establish these results, we introduce a novel analysis framework for adaptive decoding dynamics in DLMs that may be of independent interest.
Overall, our results provide the first theoretical justification for the effectiveness of confidence-based unmasking strategies, and offer a principled explanation of when and why entropy-based decoding can achieve accelerated sampling without sacrificing generation fidelity.
Our theoretical insights may inform the design of more efficient inference schemes in practice.


\subsection{Other related work}
\label{sec:related-work}

\paragraph{Convergence theory for uniform decoding strategies.}
Existing sampling convergence analysis for DLMs largely focuses on uniform decoding strategies and two main viewpoints have been pursued.
The first line of work directly studies the sampling error accumulated in the iterative unmasking procedure under the factorized approximation.
For balanced step-size schedules (i.e., $s_t\asymp L/T$ for all $t$), \citet{li2025breaking} established an $O(1/T)$ convergence rate in KL divergence, with leading factor equal to the sum of mutual information $\sum_{i=1}^T I(X^i;X^{\smallsetminus i})$ between the $i$-th token $X^i$ and the remaining tokens $X^{\smallsetminus i} \defn (X^j)_{j\neq i}$ under the data distribution.
They also proved matching lower bounds (up to constant factors), showing that the $O(1/T)$ decay is unimprovable for balanced step-size schedules in the worst case.
Subsequently, \citet{chen2025optimal} refined the convergence rate using the total correlation and dual total correlation of the data distribution, and developed optimal step-size schedules. These schedules can outperform balanced step-size schedules, albeit requiring either a priori estimates of these correlation measures or hyperparameter tuning.
Recently, \citet{zhao2026adaptation} showed that randomized step-size schedules can automatically adapt to intrinsic dependence structure underlying data, leading to improved iteration complexity for structured data distributions.
A parallel viewpoint analyzes discrete diffusion models from the perspective of CTMC discretization.
However, most results in this direction yield iteration complexity that scales at least linearly in the sequence length \citep{chen2024convergence,liang2025absorb,feng2025theoretical,ren2024discrete,ren2025fast,liang2026sharp}. Consequently, they fail to capture the parallel-generation regime in DLM decoding. A notable exception is \citet{dmitriev2026efficient}, which developed sharp and adaptive guarantees for MDMs, with convergence rate governed by low-complexity measures of data.
In addition, sample complexity guarantees have recently been established for DLMs \citep{wu2026transformer,zhanggeneralization,cho2026minimax}.

\paragraph{Empirical confidence-based strategies.}
Early confidence-based approaches typically use a fixed-size strategy, which prescribes an unmasking size $K$ and unmasks the top-$K$ most confident tokens at each iteration \citep{chang2022maskgit}.
This idea has been instantiated with several ranking criteria, including 
top-$K$ probability \citep{nie2025large}, top-$K$ margin \citep{kim2025train}, and top-$K$ entropy \citep{ben2025accelerated}.
More recent work considers adaptive strategies, where the number of newly unmasked tokens is determined on the fly.
Examples include threshold-based rules that reveal tokens whose confidence exceeds a prescribed level \citep{yu2025dimple}, and budget-based strategies that continue unmasking while the accumulated uncertainty remains within a budget \citep{ben2025accelerated,wu2025fast}. The methods analyzed in this paper fall into this category.
Beyond those one-step confidence rules, several works combine confidence information with more structured decoding procedures. For instance, 
\citet{wei2025accelerating} proposes a two-stage method that alternates between exploratory and accelerated decoding strategies depending on the confidence region.
In addition, another line of work seeks to improve decoding by explicitly planning ahead for future iterations rather than making purely myopic decisions, including lookahead-based exploration methods \citep{fu2025bits,lee2025lookahead,peng2025path} and trial-and-verification procedures \citep{wu2025free,li2026plan}.
Finally, to mitigate the limitation of standard MDMs that fix tokens once they are unmasked, recent work has also studied confidence-based remasking mechanisms that allow low-confidence tokens to be revised in later iterations
\citep{wang2025remasking,shi2024simplified}.

\paragraph{Theory for continuous diffusion models.}

A rich body of work has developed sampling convergence guarantees for continuous diffusion models in $\bR^d$ \citep{lee2022convergence,chen2022sampling,lee2023convergence,chen2023improved,benton2023linear,li2024d,li2024sharp}, with 
state-of-the-art iteration complexity of $\wt{O}(\sqrt{d/\veps})$ in KL divergence and $\wt{O}(\sqrt{d}/\veps)$ in total variation distance \citep{jiao2025optimal,zhang2025sublinear}. 
A variety of provably accelerated samplers have been further developed \citep{li2024provable,li2024improved,gatmiry2026high,huang2024convergence,li2024accelerating,li2025faster,jiao2025first}.
A theme relevant to our work is the adaptivity to (unknown) low-complexity structure. Several recent works \citep{li2024adapting,liang2025low,huang2024denoising,potaptchik2024linear,li2025dimension} show that when the data distribution concentrates near a $k$-dimensional structure with $k \ll d$, the dependence on the ambient dimension $d$ can be replaced by the intrinsic dimension $k$, yielding substantial sampling acceleration.
Beyond sampling convergence, end-to-end statistical guarantees have been developed to demonstrate the optimality of diffusion-based distribution learning and sample generation \citep{oko2023diffusion,wibisono2024optimal,zhang2024minimax,cai2025minimax,gatmiry2024learning,chen2024learning}.
Further results tailored to structured data have also been established, including distributions supported on low-dimensional linear subspaces or manifolds  \citep{chen2023score,tang2024adaptivity,azangulov2024convergence,wang2024diffusion,yakovlev2025generalization,wu2026diffusion}, distributions with low-complexity dependence \citep{fan2025optimal}, and factor models \citep{chen2025diffusion}.


\subsection{Notation}
\label{sec:notation}
For any integer $n>0$, let $[n] \defn \{1,2,\dots,n\}$.
For any set $\cS$, we use the upper-case letter $S$ to denote its cardinality $|\cS|$. 
We denote by $\bX$ the (discrete) vocabulary of tokens. We write $\mask$ for the mask and extend the vocabulary by adding $\mask$ to define $\ol \bX = \bX \cup \{\mask\}$.
Given a sequence $x\in\bX^L$, we denote its  $i$-th coordinate by $x^{i}$ for $i\in[L]$. For any index set $\cS\subseteq[L]$, let $x^\cS = (x_i)_{i\in \cS}$ be the subsequence in $\bX^{S}$ formed by the entries of $x$ with indices in $\cS$. 
%
%
Define the projection operator $\cP_\cS:\bX^L \to \ol\bX^L$ by
\begin{align}\label{eq:projection}
    \big[\cP_\cS(x)\big]^i =
    \begin{cases}
        x^i, & i\in \cS, \\
        \mask, & i\notin \cS.
    \end{cases}
\end{align}

For a random variable $X$, we use $p_X$ to denote its distribution, and (by a slight abuse of notation) also its probability mass/density function.
The entropy of $X$ is defined by $H(X) = H(p_X) \defn \bE_{x\sim p_{X}}[-\log p_X(x)]$.
For a random vector $(X,Y)\sim p_{X,Y}$,  the Kullback-Leibler divergence between $p_X$ and $p_Y$ is $\KL(p_X\,\|\, p_Y)\defn \bE_{x\sim p_{X}}[\log\frac{p_X(x)}{p_Y(x)} ]$.
The conditional entropy of $X$ given $Y$ is defined as $H(X\mid Y) \defn \bE_{(x,y)\sim p_{X,Y}}[-\log p_{X\mid Y}(x\mid y)]$.
The mutual information between $X$ and $Y$ is defined as $I(X;Y)\defn \KL(p_{X,Y}\,\|\, p_X p_Y)$.
For a random vector $(X,Y,Z)\sim p_{X,Y,Z}$, the conditional mutual information between $X$ and $Y$ given $Z$ is defined as $I(X; Y\mid Z) \defn \KL(p_{X,Y,Z} \,\|\,p_{X\mid Z}p_{Y\mid Z}p_Z)$.

In addition to these standard information-theoretic measures, we also define \textit{pointwise} conditional entropy and conditional mutual information as follows:
\begin{align}
\ptH(X\mid Z) &\defn f(Z) \quad \text{with} \quad f(z) \defn H(p_{X\mid Z=z}), \label{eq:pointwise-entropy} \\
\ptI(X,Y \mid Z) &\defn g(Z) \quad \text{with} \quad g(z) \defn \KL(p_{X,Y\mid Z=z}\,\|\, p_{X\mid Z=z}p_{Y\mid Z=z}). \label{eq:pointwise-mutual-information}
\end{align}
We note that $\ptH(X\mid Z)$ and $\ptI(X,Y\mid Z)$ are random variables measurable with respect to $Z$. Moreover, taking expectation with respect to $Z\sim p_Z$ recovers the standard quantities:
\begin{align*}
\bE_{Z}[\ptH(X\mid Z)] = H(X\mid Z) \quad \text{and} \quad \bE_{Z}[\ptI(X,Y\mid Z)] = I(X;Y\mid Z).
\end{align*}

For two functions $f(n),g(n) > 0$, we use $f(n)\lesssim g(n)$ or $f(n)=O\big(g(n)\big)$ to mean $f(n)\leq Cg(n)$ for some absolute constant $C>0$. 
Similarly, we write $f(n)\gtrsim g(n)$ or $f(n)=\Omega\big(g(n)\big)$ when $f(n)\geq C'g(n)$ for some absolute constant $C'>0$.
We denote $f(n)\asymp g(n)$ or $f(n)=\Theta\big(g(n)\big)$ when $Cf(n)\leq g(n)\leq C'f(n)$ for some absolute constants $C' > C > 0$. 
The notation $\wt{O}(\cdot)$, $\wt{\Omega}(\cdot)$, and $\wt{\Theta}(\cdot)$ denote the respective bounds up to a logarithmic factor. 

\section{Preliminaries}
\label{sec:problem}

In this section, we provide an overview of the fundamental concepts underlying DLMs.

\paragraph{Forward (masking) process.}

Consider a token sequence $X_0 = (X^1_0, X^2_0, \dots, X^L_0) \in \bX^L$ sampled from the data distribution $p_\data$. 
The forward process progressively corrupts $X_0$ by replacing its tokens with the mask symbol $\mask$ until the entire sequence is masked.
Formally, let $X_{t}$ denote the partially masked sequence obtained after $t$ masking steps and let $\cM_t \defn \{i\in[L]\colon X_{t}^i = \mask\}$ denote the index set of masked tokens.
At each step $t \geq 1$, we select a masking set $\cS_t \subseteq [L]\setminus \cM_{t-1}$ and update $X_{t-1}$ to $X_{t}$ by replacing tokens at positions in the set $\cS_t$ with $\mask$:
\begin{align*}
    X_t^i = \begin{cases}
        \mask, & \text{if } i \in \cS_t, \\
        X_{t-1}^i, & \text{if } i \notin \cS_t.
    \end{cases}
\end{align*}
The masking process continues until all tokens have been masked.

\paragraph{Mask predictor training.}
DLMs are trained to reverse the above forward masking process by learning the conditional distribution of masked tokens given the unmasked ones.
A notable feature of mask predictor training is that this posterior distribution is time-independent---it depends only on the unmasked context and not on the masking step $t$ \citep{ou2024your,zheng2024masked}.
More precisely, for each token index~$i\in[L]$ and any set of unmasked positions $\cS \subseteq [L] \setminus \{i\}$, we define the true conditional marginal distribution of $X^i_0$ given  $X_0^\cS$ under the data distribution $p_\data$ as
\begin{align}\label{eq:true-conditional}
    p^{i}( x\mid z^\cS) \defn \bP \big\{X_0^i =  x\mid X_0^\cS = z^\cS\big\} \qquad \forall\,x\in \bX,\,z^\cS\in \bX^{|\cS|}.
\end{align}
The objective is to learn estimators $p^i_\theta(\cdot \mid \cdot)$ for each conditional marginal $p^i(\cdot \mid \cdot)$, thereby forming a mask predictor $p_\theta = \prod_{i=1}^L p^i_\theta$.

The mask predictor is trained by maximizing the log-likelihood of masked tokens conditioned on unmasked ones:
\begin{align} \label{eq:objective}
\max_{p_\theta = \prod_{i=1}^L p_\theta^i
} \bE_{X_0, \cS}\Bigg[\frac{1}{|\cS^\setc|}\sum_{i\in \cS^\setc} \log p^i_\theta \big(X_0^{i} \mid X_0^\cS \big)\Bigg],
\end{align}
where $X_0\sim p_{\data}$ and $\mathcal S$ follows a prescribed masking-set distribution.
This objective is equivalent to minimizing the KL divergence between the data distribution and the distribution induced by the product of learned conditional marginals over partially masked sequences. In practice, the expectation is approximated by averaging over a finite number of training samples and random masking patterns.

We remark that the product of the true conditional marginals $\prod_{i=1}^L p^i$ defined in \eqref{eq:true-conditional} is the maximizer of the training objective \eqref{eq:objective}.

\paragraph{Reverse (unmasking) process.}

Given a trained mask predictor $p_\theta$, DLMs produce a token sequence through an iterative unmasking procedure.
Starting from a fully masked sequence $Y_0 = (\mask, \mask, \ldots, \mask)\in \ol\bX^L$ with cumulative unmasking set $\cW_0 = \varnothing$, we repeat the following two steps for iteration $t = 1, 2, \dots$ until all tokens have been unmasked.
\begin{enumerate}
    \item \textbf{Select unmasking set.} 
    Choose a subset $\cD_t$ of the currently masked positions $ [L]\setminus \cW_{t-1}$ according to a \textit{decoding strategy}, and update the cumulative unmasking set $\cW_t = \cW_{t-1} \cup \cD_t$.
    \item \textbf{Sample tokens.} 
    Draw the tokens for all positions in $\cD_t$ \textit{independently} and \textit{simultaneously} from the learned mask predictor $p_{\theta} = \prod_{i=1}^L p_\theta^i$, namely,
\begin{align}
    \wh Y^i \sim p_\theta^i\big(\cdot \mid Y_{t-1}^{\cW_{t-1}}\big),\quad \forall\,i \in \cD_t.
\end{align}
    The sequence is then updated by inserting the newly sampled tokens:
    \begin{align}
    Y_t^i = \begin{cases}
            \wh Y^i, & \text{if } i \in \cD_t, \\
            Y_{t-1}^i, & \text{if } i \notin \cD_t. 
        \end{cases}
    \end{align}
\end{enumerate}
The output is the fully unmasked sequence $Y_T$, where $T$ is the (random) number of iterations.

It is worth noting that given a fixed mask predictor $p_\theta$, the sampling process of DLMs and the sampled distribution $p_{Y_T}$ are completely determined by the choice of decoding strategy, which specifies the unmasking set $\cD_t$ at each iteration $t$.


\section{Main results}
\label{sec:main-results}

In this section, we present two entropy-based decoding strategies for DLMs and establish their theoretical guarantees. We begin with the entropy sum-based strategy, which is the main algorithmic result of the paper, and then study a maximum entropy-based variant to illustrate how the choice of entropy criterion affects sampling efficiency.

\subsection{Entropy sum-based decoding strategy}
\label{sec:entropy-sum}

\begin{algorithm}[t]
\caption{Entropy sum-based decoding strategy\label{alg:entropy-based-sampling-sum}}
    \begin{algorithmic}[1]
    \State \textbf{Input:} Mask predictor $p_\theta$ and entropy threshold $\eta$
\State Initialize $k \leftarrow 0$, $t \leftarrow 0$, token sequence $Y_0 \gets (\mask, \dots, \mask)$
\State Draw a permutation $\Pi=(\Pi_1,\dots,\Pi_L)$ of $[L]$ uniformly at random
\While{$k < L$} \Comment{Repeat until all tokens are generated}
\State $t \gets t+1$ \Comment{Start iteration $t$}
    \State 
    $Y_{t} \leftarrow Y_{t-1}$, 
    $\cD_{t} \leftarrow \varnothing$,
    $S \leftarrow 0$ \Comment{Initialize current iterate, unmasking set, and entropy sum}
    \Repeat 
        \State $k \leftarrow k + 1$
        \State sample $Y^{\Pi_k}_t \sim p^{\Pi_k}_\theta(\cdot \mid Y_{t-1})$ \Comment{Unmask token with index $\Pi_k$}
        \State $\cD_{t} \leftarrow \cD_{t} \cup \{\Pi_k\}$ \Comment{Add index $\Pi_k$ to unmasking set $\cD_t$}
        \State $S \leftarrow S + \ptH\big(p^{\Pi_k}_\theta(\cdot \mid Y_{t-1}) \big)$
         \Comment{Update entropy sum}
        \Until{$S > \eta$ or $k = L$}
        \Comment{Iteration termination criteria}
\EndWhile
\State $T \leftarrow t$ \Comment{Record total number of iterations}
\State \textbf{Output:} Generated token sequence $Y_T$
\end{algorithmic}
\end{algorithm}

We first introduce the entropy sum-based decoding strategy. Its key feature is that each decoding iteration is determined by the total uncertainty, measured by entropy, accumulated over the current set of unmasked tokens. This leads to an adaptive rule that enlarges the unmasking set automatically in low-uncertainty regions and shrinks it when the mask predictor is uncertain.

\paragraph{Decoding procedure.}
At a high level, the entropy sum-based strategy scans the remaining masked positions in a random order and continues unmasking tokens until the cumulative entropy of the current set exceeds a threshold.

%
Concretely, we first draw a permutation $\Pi$ of $[L]$ uniformly at random, where $\Pi_k$ represents the index of the $k$-th token to be unmasked.
We then repeat the following decoding procedure until all tokens have been generated.
At each iteration $t\geq 1$, we initialize the current unmasking set $\cD_t  =\varnothing$ and the running entropy sum $S = 0$.
Recall that for any set $\cW$, we use the upper-case letter $W = |\cW|$ to denote its size.
Given the partially unmasked sequence $Y_{t-1}$ from the previous iteration, in which the tokens indexed by the cumulative unmasking set
$$\cW_{t-1} = \bigcup_{i=1}^{t-1} \cD_i = \big\{\Pi_1, \ldots, \Pi_{W_{t-1}}\big\}$$ 
have been generated, we continue along the permutation and examine the positions $\Pi_k$ 
for \begin{align*}
k= W_{t-1}+1,W_{t-1}+2,\dots
\end{align*}
For each such position $\Pi_k$, we compute the entropy of the corresponding conditional marginal $p^{\Pi_k}_\theta(\cdot \mid Y_{t-1})$, generate the token $Y_t^{\Pi_k}$ from this marginal, add the index $\Pi_k$ to the current unmasking set $\cD_t$, and update the running entropy sum $S$ by $\ptH\big(p^{\Pi_k}_\theta(\cdot \mid Y_{t-1})\big)$\footnote{Here, we use the notation $\ptH$ defined in \eqref{eq:pointwise-entropy} to denote the pointwise conditional entropy. It is a random variable depending on $Y_{t-1}$, as opposed to the standard conditional entropy $H$ that averages over the randomness in $Y_{t-1}$.}.

The iteration terminates once the running entropy sum $S$ exceeds a prescribed threshold~$\eta$ or all~$L$ tokens have been generated.
Equivalently, the index set of newly unmasked tokens at iteration $t$ has the form
\begin{align*}
\cD_t = \big\{\Pi_k \colon  W_{t-1} < k \leq k_t \big\},
\end{align*}
where $k_t$ is defined by
\begin{align*}
k_t \defn \min\biggl\{ i > W_{t-1}: \sum_{k = W_{t-1}+1}^i \ptH \big(p^{\Pi_k}_\theta(\cdot \mid Y_{t-1}) \big) > \eta \biggr\} \wedge L.
\end{align*}
In other words, $k_t$ is the first permutation index beyond $W_{t-1}$ such that the cumulative entropy over permutation indices $k= W_{t-1}+1, \ldots, k_t$ exceeds $\eta$, or $L$ if no such index appears before the sequence is fully unmasked.

The complete decoding procedure is summarized in Algorithm~\ref{alg:entropy-based-sampling-sum}.

\begin{remark}
The random permutation $\Pi$ is introduced solely to simplify the presentation and analysis. Equivalently, the algorithm may be viewed as selecting the next token from the remaining masked positions uniformly at random.
\end{remark}

An important observation is that, for each iteration $t$, conditioned on the permutation $\Pi$, the cumulative unmasking set $\cW_{t-1}$, and the partially unmasked sequence $Y_{t-1}$, the unmasking set $\cD_{t}$ (and thus the cumulative unmasking set $\cW_t = \cW_{t-1} \cup \cD_t$) is fully determined by the conditional entropies:
\begin{align*}
\ptH\big(p^{\Pi_k}_\theta(\cdot \mid Y_{t-1}) \big), \quad k = W_{t-1}+1, W_{t-1}+2, \dots, L.
\end{align*}
Once $\cD_{t}$ is determined, we obtain the iterate $Y_t$ by generating the tokens in $\cD_t$ independently according to the corresponding conditional distributions 
$$Y_t^{i} \sim p^i_{\theta}(\cdot \mid Y_{t-1}), \quad i\in \cD_t,$$
or equivalently,
$$Y_t^{\Pi_k} \sim p^{\Pi_k}_\theta(\cdot \mid Y_{t-1}), \quad W_{t-1}+1\leq k\leq W_{t}.$$
Therefore, the overall sampling process can be summarized by the following Markov chain:
\begin{align}
&\Pi \to \bigl\{\ptH\big(p^{\Pi_k}_\theta(\cdot \mid Y_{0}) \big)\bigr\}_{k=1}^L \to (\cD_1,\cW_1) \to Y_1 
\to \bigl\{\ptH\big(p^{\Pi_k}_\theta(\cdot \mid Y_1) \big)\bigr\}_{k=W_1+1}^L \to (\cD_2,\cW_2) \to Y_2  \notag \\
&\qquad \to  \dots
\to \bigl\{\ptH\big(p^{\Pi_k}_\theta(\cdot \mid Y_{T-1}) \big)\bigr\}_{k=W_{T-1}+1}^L \to (\cD_{T},\cW_{T}) \to Y_T, \label{eq:markov-chain-entropy-sum}
\end{align}
where $T$ denotes the (random) total number of iterations.

\subsection{Theoretical guarantee}

We now state the performance guarantee for the entropy sum-based decoding scheme (Algorithm~\ref{alg:entropy-based-sampling-sum}) in Theorem~\ref{thm:entropy-sum-sampling}. The proof sketch can be found in Section~\ref{sec:entropy-sum-analysis}.

\begin{theorem}\label{thm:entropy-sum-sampling}
Assume that the mask predictor $p_\theta$ is optimal, namely, $p_\theta = \prod_{i=1}^L p^i$ is the product of true conditional marginals.
For any $\veps>0$, set the threshold to $\eta=\veps/\big(4(\log_2 L + 1)\big)$. Then the entropy sum-based strategy in Algorithm~\ref{alg:entropy-based-sampling-sum} achieves $\veps$-accurate sampling in KL divergence, i.e., $\bE_\Pi \big[\KL( p_{X_0} \parallel p_{Y_{T}\mid \Pi}) \big]  \leq \veps.$
Moreover, the expected number of iterations satisfies
\begin{align}
\bE[T] \leq 4\bigg(\frac{H(X_0)}{\veps}+1\bigg)(\log_2 L + 1)+1,
\end{align}
where $H(X_0)$ denotes the entropy of the target data distribution.
\end{theorem}

To the best of our knowledge, this result provides the first theoretical guarantee for confidence-based decoding strategies in DLMs.
We discuss several important implications below.
\begin{itemize}
    \item \textit{Iteration complexity.} The expected number of iterations required to achieve $\veps$-accuracy in KL divergence is at most 
    \begin{align}\label{eq:iteration-complexity-entropy-sum}
    \wt O\bigg(\frac{H(X_0)}{\veps}\bigg).
    \end{align}
    Up to a logarithmic dependence on the sequence length $L$, this iteration complexity depends only on the entropy of the target data distribution, which can be viewed as a measure of its intrinsic complexity.
    Consequently, when the target distribution has low entropy, namely $H(X_0) \ll L$, the entropy sum-based decoding scheme yields a substantial sampling speedup.
    \item \textit{Adaptivity to intrinsic structure.} The above sampling acceleration is achieved without requiring any prior knowledge of the data distribution or hyperparameter tuning. This highlights the distribution-agnostic adaptivity of the entropy-based decoding---it automatically exploits low-complexity structure in the target distribution to accelerate generation.
\end{itemize}

\subsection{Connection to empirical entropy sum-based strategies} 
Algorithm~\ref{alg:entropy-based-sampling-sum} is closely related to the entropy-based decoding strategy recently proposed in \citet{ben2025accelerated}. Their method sorts the remaining tokens by conditional entropy and selects the largest subset for which the gap between the entropy sum and the maximum entropy remains below a prescribed threshold.
Numerical results in \citet{ben2025accelerated} demonstrate that this strategy achieves substantial generation speedups while matching or surpassing the generation quality of uniform unmasking schemes.
The main difference is that Algorithm~\ref{alg:entropy-based-sampling-sum} scans token positions in a uniformly random order instead of a deterministic sorted order.
%


Despite its empirical success, entropy sum-based decoding with sorted ordering can incur large KL sampling error even for a low-entropy target distribution. The following proposition formalizes this limitation; its proof can be found in Appendix~\ref{sec:hard-instance-sorted-ordering}.
\begin{proposition}\label{prop:hard-instance-sorted-ordering}
For any entropy budget $H$ and threshold $\eta$ satisfying $H=O(\eta L),$
there exists a distribution over $\{0,1\}^L$ with entropy at most $H$ such that the entropy-sum decoding strategy with sorted ordering requires $\Omega(H/\eta)$ iterations and incurs KL sampling error $\Omega(H).$
\end{proposition}

Therefore, we adopt a uniformly random ordering to make the adaptive decoding process analytically tractable and enables us to establish nontrivial KL sampling error guarantees. 
Our main objective is to provide a theoretical justification for a central component of confidence-based decoding: adaptive set-size selection where the entropy budget controls how many tokens are decoded at each iteration. Determining under what conditions similar guarantees can be established for deterministic sorted ordering remains an interesting direction for future work.

\subsection{Effect of entropy criterion}
\label{sec:max-entropy}

A natural question is how the choice of entropy criterion affects decoding efficiency. 
To investigate this, we present a second entropy-based decoding strategy---the \textit{maximum entropy-based strategy}---which differs from the entropy sum-based strategy in a single design choice. 
Instead of controlling the cumulative entropy of the entire set, it terminates an iteration as soon as any single token's conditional entropy exceeds a prescribed threshold, subject to an explicit cap on the set size.

As we will see shortly, this strategy still achieves an iteration complexity sublinear in the sequence length. However, its sampling guarantee is weaker in two respects: the iteration complexity bound is looser, and the parameter choice requires knowledge of the target distribution. 
This contrast highlights the importance of how entropy is incorporated into the design of decoding strategies.

\paragraph{Maximum entropy-based decoding strategy.}


%

As before, we begin by drawing a random permutation~$\Pi$ of $[L]$.
At iteration $t\geq 1$, given the previous iterate $Y_{t-1}$ with cumulative unmasking set $\cW_{t-1}$, we scan the remaining positions in the permutation order and keep adding them to the current set $\cD_t$.
The iteration ends as soon as one of the following occurs:
\begin{enumerate}[label=(\roman*)]
    \item the entropy of the most recently unmasked token
     exceeds the threshold $\eta$ (this token is included in the set $\cD_t$);
    \item the current set reaches the cap size $S_{\max}$;
    \item all $L$ tokens have been generated.
\end{enumerate}
Formally, the index set $\cD_t$ of newly unmasked tokens at iteration $t$ is given by
\begin{align*}
\cD_t = \big\{\Pi_k \colon  W_{t-1} < k \leq k_t \big\},
\end{align*}
where $k_t$ is the first permutation index after $W_{t-1}$ for which the entropy exceeds $\eta$, or $W_{t-1}+S_{\max}$ if no such index appears within the next $S_{\max}$ positions, or $L$ if no such index appears before all tokens are generated. In other words, $k_t$ is defined by
\begin{align*}
k_t \defn \min\bigl\{i > W_{t-1}: \ptH\big(p^{\Pi_i}_\theta(\cdot \mid Y_{t-1}) \big) > \eta \bigr\} \wedge (W_{t-1}+S_{\max}) \wedge L.
\end{align*}

A complete description of the unmasking scheme is given in Algorithm~\ref{alg:entropy-based-sampling-max}.
%


Compared with the entropy sum-based decoding scheme, this strategy differs only in the stopping criterion:
(1) it reacts to the entropy of a single token rather than the cumulative entropy of the whole set, and (2) it explicitly imposes a maximum unmasking size $S_{\max}$.
Despite this difference, these two entropy-based strategies share the same Markovian structure: the unmasking sets $\cD_t$ and $\cW_t$ at iteration $t$ are fully determined by the previous iterate $Y_{t-1}$ and cumulative unmasking set $\cW_{t-1}$. 

\begin{algorithm}[t]
\caption{Maximum entropy-based decoding strategy\label{alg:entropy-based-sampling-max}}
\begin{algorithmic}[1]
\State \textbf{Input:} Mask predictor $p_\theta$, entropy threshold $\eta$, and maximum unmasking size $S_{\max}$
\State Initialize $k \leftarrow 0$, $t \leftarrow 0$, and token sequence $Y_0 \gets (\mask, \dots, \mask)$
\State Draw a permutation $\Pi=(\Pi_1,\dots,\Pi_L)$ of $[L]$ uniformly at random
\While{$k < L$} \Comment{Repeat until all tokens are generated}
\State $t \gets t+1$ \Comment{Start iteration $t$}
    \State 
    $Y_{t} \leftarrow Y_{t-1}$, 
    $\cD_{t} \leftarrow \varnothing$ \Comment{Initialize current iterate and unmasking set}
    \Repeat 
        \State $k \leftarrow k + 1$
        \State $Y^{\Pi_k}_t \sim p^{\Pi_k}_\theta(\cdot \mid Y_{t-1})$ \Comment{Unmask token with index $\Pi_k$}
        \State $\cD_{t} \leftarrow \cD_{t} \cup \{\Pi_k\}$ \Comment{Add index $\Pi_k$ to unmasking set $\cD_t$}
        
        \Until{$\ptH\big(p^{\Pi_k}_\theta(\cdot \mid Y_{t-1}) \big) > \eta$, or $|\cD_{t}| = S_{\max}$, or $k = L$}
        \Comment{Iteration termination criteria}
\EndWhile
\State $T \leftarrow t$ \Comment{Record total number of iterations}
\State \textbf{Output:} Generated token sequence $Y_T$
\end{algorithmic}
\end{algorithm}

\paragraph{Theoretical guarantee.}

We now present the performance guarantee of the maximum entropy-based inference scheme (Algorithm~\ref{alg:entropy-based-sampling-max}) in Theorem~\ref{thm:maximum} below. The proof can be found in Appendix~\ref{sec:max-entropy-analysis}.
\begin{theorem}\label{thm:maximum}Assume that the mask predictor $p_\theta$ is optimal, namely, $p_\theta = \prod_{i=1}^L p^i$ is the product of true conditional marginals.
For any $\veps>0$, set the entropy threshold to $\eta = \sqrt{{\veps H(X_0)}/{L}}$ and the maximum unmasking size to $S_{\max} = \lfloor \sqrt{{\veps L}/{H(X_0)}}  \rfloor \vee 1$. Then the maximum entropy-based unmasking strategy in Algorithm~\ref{alg:entropy-based-sampling-max} achieves $\veps$-accurate sampling in KL divergence, namely,
$
\bE_{\Pi}\big[\KL(p_{X_0} \parallel p_{Y_{T}\mid \Pi})\big] \leq \veps.
$
Moreover, the expected number of iterations satisfies
\begin{align}\label{eq:iteration-complexity-maximum}
\bE[T] \leq 2\sqrt{\frac{H(X_0)L}{\veps}}+1.
\end{align}
\end{theorem}

%
In words, the maximum entropy-based decoding strategy attains $\veps$-accuracy in KL divergence with an expected number of iterations scaling as
    $O\big(\sqrt{{H(X_0)L}/{\veps}}\big).$ 
    As a consequence, when the entropy of the target data distribution satisfies $H(X_0) \ll L$, it achieves a sublinear iteration complexity in the sequence length $L$, thereby implying a provable speedup over fully sequential generation.

Nevertheless, compared with the entropy sum-based strategy, this maximum entropy-based strategy has two important shortcomings.
    First, in the parallel sampling regime, where the iteration complexity is smaller than the sequence length $L$,  the bound in \eqref{eq:iteration-complexity-maximum} is worse than that in \eqref{eq:iteration-complexity-entropy-sum} by a factor of $\sqrt{\veps L /H(X_0)}\gg 1$.
    Second, the parameter choices $\eta$ and $S_{\max}$ needed to achieve a sublinear iteration complexity require explicit knowledge of the entropy $H(X_0)$ of the target data distribution. This stands in stark contrast to the entropy sum-based strategy, which is fully distribution-adaptive.

Since the theoretical guarantees for these two entropy-based decoding strategies are established under the same analysis framework, 
comparing them suggests that the choice of entropy criterion plays a critical role in efficient entropy-based decoding. 
An important future direction is thus to develop a deeper understanding of how different entropy criteria affect the sampling performance of DLMs.

\section{Analysis}
\label{sec:analysis}

In this section, we outline the proof strategy for the main theory (Theorem~\ref{thm:entropy-sum-sampling}) for the entropy sum-based decoding strategy.

\subsection{Preliminary facts}
\label{sec:preliminary-facts}

Before presenting the proof sketch, we collect several basic facts and notational conventions that will be used throughout the analysis.

\paragraph{Iterate representation.} Starting from the fully masked initialization $Y_0=(\mask,\dots,\mask)$, the iterate $Y_t$ at step $t$ is completely characterized by the cumulative unmasking set $\cW_t$ together with the revealed token values on that set, namely $Y^{\cW_t}_t$, since $Y^{i}_t = \mask$ for all $i\in \cW_t^{\setc}$.
Thus, the entropy-based sampler can be equivalently viewed as iteratively filling in the coordinates indexed by $\cW_t$. 
To simplify notation, we suppress the time index $t$ in the iterate $Y_t$ whenever  it is clear from context, and we also write $X = X_0$.

\paragraph{Permutation-induced order.} Given a fixed permutation $\Pi$ of $[L]$, the sampler scans token indices in the order $\Pi_1, \Pi_2, \ldots, \Pi_L$, where $\Pi_k$ denotes the index of the $k$-th token to be unmasked. Let $\Pi^{-1}$ denote the inverse permutation so that $\Pi^{-1}_i = k$ gives the unmasking order $k$ of the token with index $i$.
Moreover, for each token index $i\in[L]$, there exists a unique iteration $t_i$ at which token $i$ is unmasked, that is, the iteration satisfying $i \in \cD_{t_i} =\cW_{t_i} \setminus \cW_{t_i-1}$.

\paragraph{Unmasking sets.} 
The entropy-based sampling procedure produces an increasing sequence of cumulative unmasking sets $\varnothing = \cW_0 \subseteq \cW_1 \subseteq \cdots \subseteq \cW_T = [L]$ where $T$ is the random total number of iterations. For convenience of analysis, we extend this sequence by defining $\cW_t \defn [L]$ for $t > T$ so that the sequence $(\cW_t)_{t=0}^L$ is defined for all $0\leq t \leq L$.
The incremental unmasking sets $(\cD_t)_{t=1}^L$ are extended in an analogous manner by setting $\cD_t \defn \varnothing$ for $t > T$.

In addition, since tokens are revealed according to the permutation order, the cumulative unmasking set at iteration $t$ can be expressed as 
\begin{align}
\cW_t = \{\Pi_1, \ldots, \Pi_{W_t}\},
\end{align}
where we recall the notation $W_t \defn |\cW_t|$.
Equivalently, the unmasking set $\cW_t$ consists of the first $W_t$ indices in the permutation $\Pi$. 
Similarly, the set of newly unmasked tokens at iteration $t$ can be written as
\begin{align} \label{eq:unmasking-set-representation}
\cD_t = \cW_t \setminus \cW_{t-1} = \{\Pi_{W_{t-1}+1}, \ldots, \Pi_{W_t}\}.
\end{align}

\paragraph{Dependence structure of sampling process.}
Finally, we record two critical consequences of the Markovian structure of the entropy sum-based sampling procedure as shown in  \eqref{eq:markov-chain-entropy-sum}. 
%
\begin{itemize}
    \item 
First, for a fixed permutation $\Pi$, the unmasking set $\cD_t$ at iteration $t$, and hence the cumulative unmasking set $\cW_t$, are completely determined by the cumulative set $\cW_{t-1}$ and sampled token values $Y^{\cW_{t-1}}=(Y^{\Pi_1},\dots, Y^{\Pi_{|\cW_{t-1}|}})$ from the previous iteration.

\item Second, once the permutation $\Pi$ and the fully sampled token sequence $Y = (Y^{1}, \ldots, Y^{L})\in\bX^L$ are given, the entire sampling trajectory is uniquely determined.
In particular, one can recover the total number of iterations $T$, the cumulative unmasking sets $(\cW_t)_{t=0}^L$, and the incremental unmasking sets $(\cD_t)_{t=1}^L$. In other words, the total number of iterations $T$ and sets $(\cW_{t-1}, \cD_t)_{t=1}^L$ are deterministic functions of the sampled token sequence $Y$ and permutation~$\Pi$, i.e.,
\begin{align}
T = T(Y,\Pi) \qquad \text{and}\qquad  \cW_{t-1} = \cW_{t-1}(Y,\Pi), \quad \cD_t = \cD_t(Y,\Pi), \quad t\in[L]. \label{eq:unmasking-sets-deterministic}
\end{align}

\end{itemize}

\subsection{Proof sketch of Theorem~\ref{thm:entropy-sum-sampling}}
\label{sec:entropy-sum-analysis}

Having introduced the necessary preliminaries, we now outline the proof of Theorem~\ref{thm:entropy-sum-sampling} for the entropy sum-based decoding strategy.

\paragraph{Step 1: Express KL error along unmasking trajectory.}

We begin by expressing both the sampled distribution and the data distribution along the adaptive unmasking trajectory.
The key insight is that, by the characterization in \eqref{eq:unmasking-sets-deterministic}, once the permutation $\pi$ and the sampled sequence $x$ are given, the realized total iteration count $T$ and the realized sets $(w_{t-1}, d_t)_{t=1}^L$ are fully determined.

Concretely, when the mask predictor is optimal, conditioned on a fixed permutation $\Pi = \pi$, the entropy-based sampler generates the tokens in each unmasking set independently according to the true conditional marginals $p^i_\theta = p^i$ defined in \eqref{eq:true-conditional}.
Hence the density of the sampled distribution factorizes within each set as a product of conditional marginals:
\begin{subequations}\label{eq:low-dim-densities}
\begin{align}\label{eq:sample-density-low-dim}
\log p_{Y_{T}\mid \Pi = \pi}(x) & = \sum_{t=1}^{T}  \sum_{i \in d_t} \log p(x^{i}\mid x^{w_{t-1}}) = \sum_{t=1}^{L} \ind\{t\leq T\}  \sum_{i \in d_t} \log p(x^{i}\mid x^{w_{t-1}}), \quad x\in\bX^L,
\end{align}
where the total number of iterations $T = T(x,\pi)$ and the sets $d_t = d_t(x,\pi)$ and $w_{t-1} = w_{t-1}(x,\pi)$ for $1\leq t\leq L$ are uniquely determined by $x$ and $\pi$.

By contrast, under the data distribution, the tokens revealed at each iteration follow their true conditional joint distribution. 
Grouping the density according to the same unmasking sets $(w_{t-1},d_t)_{t=1}^L$ gives
\begin{align}\label{eq:data-density-low-dim}
\log p_{X_0}(x) =  \sum_{t=1}^{T} \log p(x^{d_t}\mid x^{w_{t-1}}) = \sum_{t=1}^{L} \ind\{t\leq T\}  \log p(x^{d_t}\mid x^{w_{t-1}}).
\end{align}
\end{subequations}

Using the two factorizations above, let us calculate the KL divergence between the data and sampled distributions.
Fixing a permutation $\Pi = \pi$, we can combine \eqref{eq:sample-density-low-dim} and \eqref{eq:data-density-low-dim} to derive
\begin{align}\label{eq:kl-low-dim-tower}
& \KL\big( p_{X_0} \parallel p_{Y_{T}\mid \Pi = \pi}\big) \notag\\
& \quad  = \bE_{X\sim p_{X_0}} \Bigg[ \sum_{t=1}^L  \ind\{t \leq T\} \log \frac{ p(X^{\cD_t}\mid X^{\cW_{t-1}})}{\prod_{i \in \cD_t} p(X^{i}\mid X^{\cW_{t-1}})} \,\Big|\, \Pi = \pi \Bigg]  \notag \\
& \quad = \bE \Bigg[ \sum_{t=1}^L  \ind\{t\leq T\} \bE \bigg[ \log \frac{ p(X^{\cD_t}\mid X^{\cW_{t-1}})}{\prod_{i \in \cD_t} p(X^{i}\mid X^{\cW_{t-1}})} \,\Big|\, \cW_{t-1},X^{\cW_{t-1}},\Pi = \pi \bigg] \,\big|\,\Pi = \pi \Bigg], 
\end{align}
where the last line holds due to the tower property, linearity of expectation, and the fact that $\{t\leq T\} = \{\cW_{t-1} \neq [L]\}$ can be determined by $\cW_{t-1}$.
Importantly, the only randomness here (with $\Pi=\pi$ fixed) comes from $X\sim p_{X_0}$. Once $X$ is fixed, the iteration count $T$, the sets $(\cW_{t-1},\cD_t)_{t=1}^L$ and the subsequences $(X^{\cW_{t-1}},X^{\cD_t})_{t=1}^L$ are all determined.

We next study the inner conditional expectation, which corresponds to the contribution of iteration $t$ to the overall KL sampling error.
Fix an iteration $t\in[L]$, and condition on $\Pi = \pi$ and $(\cW_{t-1},X^{\cW_{t-1}}) = (w_{t-1},x^{w_{t-1}})$. 
Under this conditioning, the set $\cD_t = d_t$ is deterministic, and we can derive
\begin{align}
& \bE \bigg[ \log \frac{ p(X^{d_t}\mid X^{w_{t-1}})}{\prod_{i \in d_t} p(X^{i}\mid X^{w_{t-1}})} \,\Big|\, (\cW_{t-1},X^{\cW_{t-1}}) = (w_{t-1},x^{w_{t-1}}),\Pi = \pi \bigg] \notag \\
& \qquad \numpf{i}{=}  \bE_{X^{d_t}} \bigg[  \log \frac{  p(X^{d_t}\mid x^{w_{t-1}})}{ \prod_{i \in d_t} p(X^{i}\mid x^{w_{t-1}})} \,\Big|\, (\cW_{t-1},X^{\cW_{t-1}}) = (w_{t-1},x^{w_{t-1}}),\Pi = \pi \bigg] \notag \\ 
& \qquad \numpf{ii}{=}  \KL \Big(p^{d_t}( \cdot \mid x^{w_{t-1}}) \,\big\|\, \prod_{i \in d_t} p^i(\cdot \mid  x^{w_{t-1}}) \Big), \label{eq:kl-low-dim-inner-1} 
\end{align}
where (i) is true because the unmasking set $\cD_t = d_t$ becomes deterministic conditioned on $\Pi = \pi$ and $(\cW_{t-1},X^{\cW_{t-1}}) = (w_{t-1},x^{w_{t-1}})$;
(ii) holds because the expectation is taken with respect to $X^{d_t}$ under the true conditional distribution $p( \cdot \mid x^{w_{t-1}})$, which is independent of $\Pi$.

Substituting \eqref{eq:kl-low-dim-inner-1} into \eqref{eq:kl-low-dim-tower} and then averaging over both $X\sim p_{X_0}$ and $\Pi$ yields
\begin{align}
\bE_{\Pi}\big[\KL(p_{X_0} \parallel p_{Y_{T}\mid \Pi})\big] 
& = \bE_{\Pi,X\sim p_{X_0}}\Bigg[ \sum_{t=1}^T \KL \Big(p^{\cD_t}( \cdot \mid X^{\cW_{t-1}}) \,\big\|\, \prod_{i \in \cD_t} p^i(\cdot \mid  X^{\cW_{t-1}}) \Big) \Bigg]. \label{eq:kl-low-dim-tower-sum-expression}
\end{align}
Thus, the expected KL sampling error decomposes as a sum over iterations $t$, where each term measures the discrepancy between the true conditional joint distribution of the tokens unmasked at iteration $t$ and the product of their conditional marginals.

\paragraph{Step 2: Decompose KL error into sum of mutual information.}

In light of the expression for the total KL error in \eqref{eq:kl-low-dim-tower-sum-expression}, it suffices to analyze the contribution from each iteration. This reduces to understanding the KL divergence between a joint distribution and the product of its marginals.

To this end, Lemma~\ref{lem:kl-mutual-info-decomp} below states that this KL divergence can be decomposed into a sum of mutual information terms, each involving one token and the tokens with larger indices.
The proof can be found in Appendix \ref{sec:proof-kl-mutual-info-decomp}.
\begin{lemma}\label{lem:kl-mutual-info-decomp}
Consider a random sequence $(Z^1, \ldots, Z^d)$ with joint distribution $p_{Z^1, \ldots, Z^d}$ and marginal distributions $p_{Z^i}$ for $i \in [d]$.
The KL divergence between the joint distribution and the product of marginals is given by
\begin{align}
\mathsf{KL}\Big(p_{Z^1, \ldots, Z^d} \,\big\|\, \prod_{i\in[d]} p_{Z^i}\Big)
& = \sum_{i = 1}^{d-1} I(Z^{i}; Z^{i+1},Z^{i+2},\dots, Z^d). \label{eq:kl-mutual-info-decomp}
\end{align}
Moreover, for any $j \in [d]$, the KL divergence can be upper bounded as
\begin{align}
\mathsf{KL}\Big(p_{Z^1, \ldots, Z^d} \,\big\|\, \prod_{i=1}^d p_{Z^i}\Big) 
& \leq       \sum_{i\in[d]\setminus \{j\}} I(Z^{i}; Z^1,\dots,Z^{i-1},Z^{i+1}\dots,Z^d). \label{eq:kl-mutual-info-decomp-2}
\end{align}
\end{lemma}
\begin{remark}
Both expressions involve a sum of $d-1$ terms, but they differ in structure. In the identity \eqref{eq:kl-mutual-info-decomp}, the last index $i = d$ is excluded, and each term measures the mutual information between $Z^i$ and the variables with larger indices $Z^{i+1},Z^{i+2},\dots, Z^d$. In the upper bound \eqref{eq:kl-mutual-info-decomp-2}, one instead excludes an arbitrary index $j\in[d]$, and each remaining term measures the mutual information between $Z^i$ and all other variables $Z^k$ with $k \neq i$.
\end{remark}

We shall apply the upper bound \eqref{eq:kl-mutual-info-decomp-2} in Lemma~\ref{lem:kl-mutual-info-decomp} to control the per-iteration KL sampling error.
The key point is that, by construction, each iteration stops as soon as the running entropy sum of the currently unmasked tokens exceeds the threshold $\eta$. Consequently, within each iteration, there can be at most one token whose entropy is larger than $\eta$. 
Motivated by this observation, define
\begin{align}
j_t & \defn \argmax_{i \in \cD_t} \ptH (X^i \mid \cW_{t-1},X^{\cW_{t-1}}),
\end{align}
with ties broken arbitrarily. Then every index $i\in \cD_t \setminus \{j_t\}$ satisfies
\begin{align*}
\ptH (X^i \mid \cW_{t-1},X^{\cW_{t-1}}) \leq \eta.
\end{align*}
Here, we remind the reader that $\ptH(X\mid Z)$ denotes the entropy of $X$ conditioned on a realization of~$Z$ (without taking the expectation of $Z$), so it is a function of~$Z$. Similarly, $\ptI(X;Y_T\mid Z)$ denotes the mutual information between $X$ and $Y$ given a  realization of~$Z$.

Now applying \eqref{eq:kl-mutual-info-decomp-2} in Lemma~\ref{lem:kl-mutual-info-decomp} with $p_{Z^1, \ldots, Z^d} = p^{D_t}( \cdot \mid x^{w_{t-1}})$ and $j = j_t$,
we obtain
\begin{align}
& \KL \Big(p^{\cD_t}(\cdot \mid X^{\cW_{t-1}}) \,\big\|\, \prod_{i \in \cD_t} p^i(\cdot \mid  X^{\cW_{t-1}}) \Big) \notag \\
& \quad \leq \sum_{i\in \cD_t} \ptI(X^{i}; X^{\cD_t\setminus \{i\}} \mid \cW_{t-1},X^{\cW_{t-1}},\Pi)\,\ind{\{i \neq j_t\}} \notag \\ 
& \quad \leq \sum_{i\in \cD_t} \ptI(X^{i}; X^{\cD_t\setminus \{i\}} \mid \cW_{t-1},X^{\cW_{t-1}},\Pi)\,\ind{\big\{\ptH (X^i \mid \cW_{t-1},X^{\cW_{t-1}}) \leq \eta\big\}}. \label{eq:kl-low-dim-decomp-entropy-sum}
\end{align}
\begin{remark}
Conditioned on $\cW_{t-1}$ and $X^{\cW_{t-1}}$, the conditional entropy
$\ptH(X^i \mid \cW_{t-1}, X^{\cW_{t-1}})$ is independent of the permutation $\Pi$, i.e.,
\begin{align*}
\ptH(X^i \mid \cW_{t-1}, X^{\cW_{t-1}}) = \ptH(X^i \mid \cW_{t-1}, X^{\cW_{t-1}}, \Pi).
\end{align*}
This is in contrast to the mutual information terms, which depend explicitly on
the unmasking set $\cD_t$, and hence on $\Pi$.
\end{remark}

Thus, the KL sampling error contributed by iteration $t$ is controlled by a sum of mutual information, where each term quantifies the dependence between one unmasked token and the rest of the tokens unmasked in iteration $t$, conditioned on the tokens revealed previously.
The indicator function in \eqref{eq:kl-low-dim-decomp-entropy-sum} highlights an important feature of the entropy-sum stopping rule: within each iteration, only tokens with conditional entropy at most $\eta$ contribute to this KL error bound.
As will be clear momentarily, this indicator function will be essential in the subsequent analysis.

Substituting the bound \eqref{eq:kl-low-dim-decomp-entropy-sum} into the KL error expression in \eqref{eq:kl-low-dim-tower-sum-expression}, we find
\begin{align}
& \bE_\Pi \Big[\KL\big( p_{X_0} \parallel p_{Y_T \mid \Pi}\big) \Big] \notag\\
&\qquad \leq \bE_{X,\Pi}\Bigg[ \sum_{t=1}^T \sum_{i\in \cD_t} \ptI(X^{i}; X^{\cD_t\setminus \{i\}} \mid \cW_{t-1},X^{\cW_{t-1}},\Pi) \, \ind{\big\{\ptH(X^i \mid \cW_{t-1},X^{\cW_{t-1}}) \leq \eta\big\}} \Bigg] \notag\\
&\qquad = \sum_{i=1}^L \bE_{X,\Pi}\Big[   \ptI(X^{i}; X^{\cD_{t_i}\setminus \{i\}} \mid \cW_{t_i-1},X^{\cW_{t_i-1}},\Pi) \, \ind{\big\{\ptH(X^i \mid \cW_{t_i-1},X^{\cW_{t_i-1}}) \leq \eta\big\}} \Big], \label{eq:kl-low-dim-tower-sum}
\end{align}
where the last step rearranges the summation order from the iteration index $t$ to the token index $i$, where we recall that $t_i$ denotes the index of the iteration at which token~$i$ is unmasked.

In summary, the expected KL sampling error is bounded by a sum of mutual information terms over token index $i\in[L]$. Each term measures the dependence between token $i$ and the other unmasked tokens in the same iteration, conditioned on the already unmasked tokens and the permutation.

\paragraph{Step 3: Average over random unmasking order.}

Equipped with the KL error bound in \eqref{eq:kl-low-dim-tower-sum}, it remains to control each token-wise contribution separately.

To this end, fix a token index $i$ and consider the corresponding summand:
\begin{align}
\bE_{X,\Pi_{\smallsetminus i},\Pi^{-1}_i}\Big[\ptI(X^{i}; X^{\cD_{t_i}\setminus \{i\}} \mid \cW_{t_i-1},X^{\cW_{t_i-1}},\Pi) \ind{\big\{\ptH(X^i \mid \cW_{t_i-1},X^{\cW_{t_i-1}}) \leq \eta\big\}}\Big]. \label{eq:kl-low-dim-tower-sum-i-1}
\end{align}
To make the dependence on the permutation explicit, we decompose the randomness of $\Pi$ into the position~$\Pi^{-1}_i$ of token $i$ in the permutation $\Pi$ and the relative ordering of all other tokens, denoted by $\Pi_{\smallsetminus i}$.
Concretely, we define
\begin{align}\label{eq:aux-permutation}
\Pi_{\smallsetminus i} \defn \big(\Pi_1, \dots, \Pi_{\Pi^{-1}_i-1}, \Pi_{\Pi^{-1}_i+1}, \dots, \Pi_L \big).
\end{align}
That is, the permutation $\Pi_{\smallsetminus i}$ is obtained from $\Pi$ by removing the index $i$ while preserving the relative order of all other tokens.

Our goal is to show, after averaging over the randomness of the position $\Pi^{-1}_i$, the contribution of token $i$ to the KL sampling error admits a clean and tractable upper bound.
However, the major technical difficulty arises from the complicated dependence between the set $\cD_{t_i}\setminus \{i\}$---the tokens revealed in the same iteration as token $i$---and the full permutation $\Pi$. 
In particular, the iteration index $t_i$ and the set $\cD_{t_i}$ both vary with $\Pi^{-1}_i$, making it challenging to directly analyze \eqref{eq:kl-low-dim-tower-sum-i-1}.

To address this, recall that $D_t \defn |\cD_t|$ denotes the size of the unmasking set $\cD_t$ at iteration $t\in[L]$. We then introduce a sequence of dyadic size envelopes $(\ol{D}_t)_{t=1}^L$ in $\bR$, defined recursively by $\ol{D}_1 = 0$ and, for~$2 \leq t \leq L$, 
\begin{align}\label{eq:size-envelope-sequence}
\ol{D}_{t} \defn 
\begin{cases}
\ol{D}_{t-1}, & \text{if}~~ D_{t-1} < \ol{D}_{t-1},\\
2 D_{t-1}, & \text{if}~~ D_{t-1} \geq \ol{D}_{t-1}.
\end{cases}
\end{align}
In essence, the envelope remains unchanged as long as the previous set size stays below it; once the previous set size reaches the envelope, the envelope is doubled.

Using the size-envelopes constructed above, the following lemma shows that, after averaging over the random position of token $i$, the mutual-information term in \eqref{eq:kl-low-dim-tower-sum-i-1} can be bounded by a much simpler quantity involving only the conditional entropy of $X^i$ and the event that the corresponding set size crosses the envelope. The proof is deferred to Appendix~\ref{sec:proof-entropy-sum-token-i-kl-bound}.
\begin{lemma}\label{lem:entropy-sum-token-i-kl-bound}
For any token index $i\in[L]$, we have
\begin{align}\label{eq:entropy-sum-token-i-kl-bound}
& \bE_{X,\Pi_{\smallsetminus i},\Pi^{-1}_i}\Big[\ptI(X^{i}; X^{\cD_{t_i}\setminus \{i\}} \mid \cW_{t_i-1},X^{\cW_{t_i-1}},\Pi) \ind{\big\{\ptH(X^i \mid \cW_{t_i-1},X^{\cW_{t_i-1}}) \leq \eta\big\}}\Big] \notag \\
& \qquad \leq 2\,\bE_{X,\Pi_{\smallsetminus i}} \Big[ \big( \ptH (X^i \mid \cW_{t_i-1},X^{\cW_{t_i-1}}) \wedge \eta \big) \ind\big\{D_{t_i} \geq \ol{D}_{t_i}\big\}  \Big].
\end{align}
\end{lemma}
In words, after averaging over the random unmasking order of token $i$, its contribution to the KL sampling error is controlled by its truncated conditional entropy, and contributes only on the event that the size of its unmasking set is at least as large as the corresponding envelope.
Thus, the intricate dependence structure in \eqref{eq:kl-low-dim-tower-sum-i-1} is reduced to a much more tractable indicator of whether a size-doubling event occurs.

\paragraph{Step 4: Control KL error by entropy-based stopping rule.}

Summing the bound \eqref{eq:entropy-sum-token-i-kl-bound} from Lemma~\ref{lem:entropy-sum-token-i-kl-bound} over all tokens and combining it with the KL error expression in \eqref{eq:kl-low-dim-tower-sum}, we obtain
\begin{align}\label{eq:entropy-sum-kl-bound-step3-1}
\bE_\Pi \big[\KL( p_{X_0} \parallel p_{Y_T\mid \Pi}) \big] 
& \leq 2\,\bE\Bigg[ \sum_{i=1}^L  \big( \ptH (X^i \mid \cW_{t_i-1},X^{\cW_{t_i-1}},\Pi_{\smallsetminus i} ) \wedge \eta \big) \ind\big\{D_{t_i} \geq \ol{D}_{t_i}\big\} \Bigg].
\end{align}
We now reindex the sum by iteration rather than token.
Since $\cD_t$ is exactly the set of tokens revealed at iteration $t$,
the right-hand side of \eqref{eq:entropy-sum-kl-bound-step3-1} becomes
\begin{align*}
& \bE\Bigg[ \sum_{i=1}^L  \big( \ptH (X^i \mid \cW_{t_i-1},X^{\cW_{t_i-1}},\Pi_{\smallsetminus i} ) \wedge \eta \big) \ind\big\{D_{t_i} \geq \ol{D}_{t_i}\big\} \Bigg] \notag\\
& \qquad = \bE\Bigg[ \sum_{t=1}^L \ind\{D_t \geq \ol{D}_{t}\} \sum_{i\in \cD_t}  \big( \ptH (X^i \mid \cW_{t-1},X^{\cW_{t-1}},\Pi_{\smallsetminus i} ) \wedge \eta \big) \Bigg] 
\end{align*}

Next, we bound the inner sum using the stopping rule. By construction, within each iteration the cumulative entropy remains below $\eta$ until the final token is added. Therefore, the total contribution of all but the last token is at most $\eta$, while the final token contributes at most $\eta$ after truncation. Consequently,
\begin{align*}
\sum_{i\in \cD_t}  \big( \ptH (X^i \mid \cW_{t-1},X^{\cW_{t-1}},\Pi_{\smallsetminus i} ) \wedge \eta \big)  \leq 2\eta,
\end{align*}
and hence
\begin{align}
\label{eq:entropy-sum-kl-bound-step3-2}
\bE\Bigg[ \sum_{i=1}^L  \big( \ptH (X^i \mid \cW_{t_i-1},X^{\cW_{t_i-1}},\Pi_{\smallsetminus i} ) \wedge \eta \big) \ind\big\{D_{t_i} \geq \ol{D}_{t_i}\big\} \Bigg] \leq 2\eta \cdot \bE\Bigg[ \sum_{t=1}^L \ind\{D_t \geq \ol{D}_{t}\} \Bigg].
\end{align}

It remains to bound the number of iterations for which the event $D_t \geq \ol{D}_{t}$ occurs.
Let $s_1 < s_2 < \dots$ be the iterations at which this event holds. By the definition of the size-envelope sequence in \eqref{eq:size-envelope-sequence}, whenever $D_{s_j} \geq \ol{D}_{s_j}$ occurs, the envelope at the next such iteration must have doubled. Thus, we have $$D_{s_{j+1}} \geq \ol{D}_{s_{j+1}} = 2 D_{s_{j}},$$ and by induction, $$D_{s_j} \geq 2^{j-1} D_{s_1} = 2^{j-1}.$$ Since $D_{s_j}\leq L$ for all $j$, this implies that the number of such iterations satisfies
\begin{align}\label{eq:entropy-sum-kl-bound-step3-3}
    \sum_{t=1}^L \ind\{D_t \geq \ol{D}_{t}\} \leq \log_2 L + 1.
\end{align}

Finally, combining \eqref{eq:entropy-sum-kl-bound-step3-1}--\eqref{eq:entropy-sum-kl-bound-step3-3} gives the final bound on the expected KL sampling error:
\begin{align}\label{eq:entropy-sum-kl-bound-step3-4}
\bE_\Pi \big[\KL( p_{X_0} \parallel p_{Y_T\mid \Pi}) \big]  & \leq 4\eta (\log_2 L + 1).
\end{align}
Setting $\eta = \veps / (4(\log_2 L + 1))$ immediately yields the desired result $\bE_\Pi \big[\KL( p_{X_0} \parallel p_{Y_T\mid \Pi}) \big] \leq \veps$.

\paragraph{Step 5: Bound expected number of iterations.}
Finally, let us study the expected number of iterations $\bE[T]$.
To this end, we use the following entropy identity, whose proof is deferred to Appendix~\ref{sec:lem:entropy-expression}.
\begin{lemma}\label{lem:entropy-expression}
For any fixed permutation $\Pi$, the entropy of the data distribution admits the following decomposition:
\begin{align}
H(X_0) &= \bE_{X\sim p_{X_0}}\Bigg[\sum_{t=1}^T \ptH(X^{\cD_t} \mid \cW_{t-1}, X^{\cW_{t-1}}, \Pi)\Bigg] \label{eq:entropy-expression-joint} \\
& = \bE_{X\sim p_{X_0}}\Bigg[\sum_{t=1}^T \sum_{i\in \cD_t} \ptH(X^{i} \mid \cW_{t-1}, X^{\cW_{t-1}}, \Pi) \Bigg] - \KL(p_{X_0} \parallel p_{Y_T\mid \Pi}). \label{eq:entropy-expression-margin}
\end{align}
\end{lemma}

We now combine this identity with the stopping rule. By construction, every iteration except the last one accumulates conditional entropy at least $\eta$, that is,
\begin{align*}
\sum_{i\in \cD_t} \ptH(X^{i} \mid \cW_{t-1}, X^{\cW_{t-1}}, \Pi) \geq \eta,
\end{align*}
for all $t < T$.
Therefore, substituting this into the entropy identity \eqref{eq:entropy-expression-margin} yields
\begin{align*}
H(X_0) \geq \bE[T-1]\eta - \bE_{\Pi}\big[\KL(p_{X_0} \parallel p_{Y_T\mid \Pi})\big].
\end{align*}
Using the KL bound from \eqref{eq:entropy-sum-kl-bound-step3-4}, we further obtain
\begin{align*}
H(X_0) \geq \big(\bE[T]-1-4(\log_2 L + 1)\big)\eta.
\end{align*}

Therefore, rearranging the above inequality and substituting $\eta = \veps / (4(\log_2 L + 1))$ into it yields
\begin{align*}
\bE[T] \leq \frac{H(X_0)}{\eta} + 4(\log_2 L + 1) + 1 = 4\bigg(\frac{H(X_0)}{\veps}+1\bigg)(\log_2 L + 1) + 1.
\end{align*}

This completes the proof of Theorem~\ref{thm:entropy-sum-sampling}.


\section{Discussion}\label{sec:discussion}

In this work, we have made progress toward understanding the sampling efficiency of confidence-based decoding strategies in DLMs. Our results show that entropy-based unmasking strategies can provably accelerate sampling, achieving an iteration complexity that is sublinear in the sequence length when the data distribution has low entropy. Notably, they reveal that entropy-based strategies can adapt to the unknown intrinsic complexity of the data distribution without requiring any prior knowledge or hyperparameter tuning.

Moving forward, it would be of interest to understand the performance of the entropy-based decoding schemes based on a deterministic order (e.g., a sorted order) instead of the random order considered in this work. Also, it would be valuable to extend our analysis to more general confidence measures beyond entropy, such as the mass of the most likely token and the mass gap between the two most likely tokens.
Second, our current analysis focuses on the scenario where the optimal mask predictor is used. It would be important to understand the impact of training error in the mask predictor on the sampling performance under confidence-based decoding, which may yield insights into the design of more effective training objectives for DLMs.
Third, another important direction is to establish lower bounds on the iteration complexity of confidence-based decoding schemes, which would shed light on the fundamental limits of sampling acceleration when using this class of strategies.

\section*{Acknowledgements}

C.~Cai is supported in part by the NSF CAREER Award CCF-2541600 and the NSF grant DMS-2515333.
G.\ Li is supported in part by the Chinese University of Hong Kong Direct Grant for Research and the Hong Kong Research Grants Council ECS 2191363, GRF 2131005.

\bibliographystyle{apalike}
\bibliography{bibfileDF,bibfileLLM}

\appendix
\section{Proof of lemmas for Theorem~\ref{thm:entropy-sum-sampling}}
\label{sec:proof-lemma-sum}

\subsection{Proof of Lemma~\ref{lem:kl-mutual-info-decomp}}
\label{sec:proof-kl-mutual-info-decomp}

To simplify the notation, we write $Z^{>i} = (Z^{i+1}, \ldots, Z^d)$ for any $1\leq i < d$ and define $Z^{<i}$, $Z^{\geq i}$, and $Z^{\leq i}$ analogously. In addition, we denote $Z^{\smallsetminus i} \defn (Z^1,\dots, Z^{i-1},Z^{i+1}, \ldots, Z^d)$ for any $1\leq i \leq d$.
Using the chain rule, one can derive
\begin{align*}
\mathsf{KL}\Big(p_{Z^1, \ldots, Z^d} \,\big\|\, \prod_{i=1}^d p_{Z^i}\Big)
& = \int p_{Z^1, \ldots, Z^d}(z^1,\ldots,z^d) \log\frac{\prod_{i=1}^d p_{Z^i \mid Z^{>i}}(z^i \mid z^{>i})}{\prod_{i=1}^d p_{Z^i}(z^i)} \diff z^1 \ldots \diff z^d \notag\\
& = \sum_{i = 1}^{d-1} \int p_{Z^{<i} \mid Z^{\geq i}}(z^{<i} \mid z^{\geq i}) \diff z^{<i} \int p_{Z^{\geq i}} (z^{\geq i}) \log\frac{p_{Z^i \mid Z^{>i}}(z^i \mid z^{>i})}{p_{Z^i}(z^i)} \diff z^{\geq i} \notag\\
& = \sum_{i = 1}^{d-1} \int p_{Z^i,Z^{>i}}(z^i,z^{>i}) \log\frac{p_{Z^i \mid Z^{>i}}(z^i \mid z^{>i})}{p_{Z^{i}}(z^{i})} \diff z^i \diff z^{>i} \notag\\
& = \sum_{i = 1}^{d-1} I(Z^{i}; Z^{>i}),
\end{align*}
where the second step holds because $p_{Z^d \mid Z^{>d}}(z^d \mid z^{>d}) = p_{Z^d}(z^d)$.
This proves \eqref{eq:kl-mutual-info-decomp}.

Turning to \eqref{eq:kl-mutual-info-decomp-2}, let us fix an arbitrary $j\in[d]$. We shall apply the same identity as in \eqref{eq:kl-mutual-info-decomp} but choose an ordering of the random variables such that $Z^j$ placed last. Define
\begin{align*}
(Y^1,\dots, Y^d) \defn (Z^1,\dots, Z^{j-1}, Z^{j+1}, \ldots, Z^d, Z^j).
\end{align*} 
Recognizing that the KL divergence is invariant to the ordering of the random variables, one has
\begin{align*}
\KL(p_{Z^1, \ldots, Z^d} \,\big\|\, \prod_{i=1}^d p_{Z^i}) & = \KL(p_{Y^1, \ldots, Y^d} \,\big\|\, \prod_{i=1}^d p_{Y^i}) = \sum_{k=1}^{d-1} I(Y^k; Y^{>k}),
\end{align*}
where the last step applies \eqref{eq:kl-mutual-info-decomp} to the random sequence $(Y^1,\dots, Y^d)$.

Note that for each $1\leq k < d$, $Y^k$ is one of the random variables $Z^i$ with $i \neq j$ and $Y^{>k} \subseteq Z^{\smallsetminus i}$. Therefore, by the monotonicity of mutual information, one has $$I(Y^k; Y^{>k}) \leq I(Z^i; Z^{\smallsetminus i}).$$
Summing over all $i\neq j$ gives the desired bound in \eqref{eq:kl-mutual-info-decomp-2}:
\begin{align*}
\KL(p_{Z^1, \ldots, Z^d} \,\big\|\, \prod_{i=1}^d p_{Z^i}) \leq \sum_{i\neq j} I(Z^i; Z^{\smallsetminus i}).
\end{align*}
This completes the proof of Lemma~\ref{lem:kl-mutual-info-decomp}.

\subsection{Proof of Lemma~\ref{lem:entropy-sum-token-i-kl-bound}}
\label{sec:proof-entropy-sum-token-i-kl-bound}

\paragraph{Preparation: auxiliary sequence.}
Before presenting the details, let us first make several preparations to address the statistical coupling issue mentioned in Section~\ref{sec:analysis}.

Consider the auxiliary sampling process where we want to unmask the token sequence except the $i$-th token, i.e., 
$$X^{\smallsetminus i} = (X^{1},\dots,X^{i-1},X^{i+1},\dots,X^{L}),$$ and run the entropy sum-based sampling procedure (Algorithm~\ref{alg:entropy-based-sampling-sum}) according to the order induced by the auxiliary permutation in \eqref{eq:aux-permutation}:
\begin{align*}
\Pi_{\smallsetminus i} \defn \big(\Pi_1, \dots, \Pi_{\Pi^{-1}_i-1}, \Pi_{\Pi^{-1}_i+1}, \dots, \Pi_L \big).
\end{align*}
For each $t\in[L]$, let $\cW^{\smallsetminus i}_t$ and $\cD^{\smallsetminus i}_t$ denote the cumulative and incremental unmasking sets at iteration $t$ in this auxiliary sampling process, respectively.
In addition, we denote their sizes by $W^{\smallsetminus i}_t \defn |\cW^{\smallsetminus i}_t|$ and
$D^{\smallsetminus i}_t \defn |\cD^{\smallsetminus i}_t|$.

Moreover, similar to the size envelopes $(\ol{D}_t)_{t=1}^L$ in \eqref{eq:size-envelope-sequence}, we define the size envelopes associated with the auxiliary process $(\ol{D}_t^{\smallsetminus i})_{t=1}^L$ recursively by $\ol{D}^{\smallsetminus i}_1 \defn 0$, and for~$t\geq 2$, 
\begin{align}\label{eq:size-envelope-sequence-aux}
\ol{D}^{\smallsetminus i}_{t} \defn 
\begin{cases}
\ol{D}^{\smallsetminus i}_{t-1}, & \text{if}~~ D^{\smallsetminus i}_{t-1} < \ol{D}^{\smallsetminus i}_{t-1},\\
2 D^{\smallsetminus i}_{t-1}, & \text{if}~~ D^{\smallsetminus i}_{t-1} \geq \ol{D}^{\smallsetminus i}_{t-1}.
\end{cases}
\end{align}

Below we make several important observations about the auxiliary sampling process.
\begin{enumerate}[label=\arabic*.,ref={\thesection\arabic*}]
\item By construction, the auxiliary unmasking sets $(\cD^{\smallsetminus i}_t)_{t\geq 1}$ are produced by running Algorithm~\ref{alg:entropy-based-sampling-sum} using the auxiliary permutation $\Pi_{\smallsetminus i}$. Hence, for any iteration $t \geq 1$, the set $\cD^{\smallsetminus i}_t$ is independent of the unmasking order~$\Pi^{-1}_i$ of token $i$ in the original process.
\item For any iteration $t\geq 1$, conditioned on $\Pi_{\smallsetminus i}$ and $(\cW^{\smallsetminus i}_{t-1},X^{\cW^{\smallsetminus i}_{t-1}})$, the unmasking set $\cD^{\smallsetminus i}_{t}$ at iteration $t$ in the auxiliary process is deterministic. 
\item \label{obs:coincide}
Recall that $t_i$ denotes the iteration at which the token with index $i$ is unmasked in the original sampling process.
Note that before this iteration, the original sampling process never uses the information contained in the $i$-th token to determine unmasking sets and sample tokens. 
This implies that the unmasking sets and sampled tokens before iteration $t_i$ coincide between the original and auxiliary processes.
In particular, $\cW^{\smallsetminus i}_{t} = \cW_{t}$ and $\cD^{\smallsetminus i}_{t} = \cD_{t}$ for all $t < t_i$ in both processes.

\end{enumerate}

To proceed, we make explicit some statistical dependence.

\begin{itemize}
    \item After conditioning on $\Pi_{\smallsetminus i}$ and $X^{\smallsetminus i}$, the only remaining randomness in the original sampling process that affects the iteration at which token $i$ is unmasked is the insertion position $\Pi^{-1}_i$ of token $i$ in the permutation $\Pi$.
To highlight this, we write
\begin{align*}
t_i = t_i(\Pi^{-1}_i)
\end{align*}
for the iteration at which token $i$ is unmasked when its insertion position is $\Pi^{-1}_i$.

\item
More generally, in the original sampling process, the unmasking set $\cD_{t_i}$ containing token $i$ depends on $\Pi^{-1}_i$ through both the iteration index and the order of token $i$ within that iteration. Hence, we write
\begin{align}\label{eq:unmasking-set-dependence}
\cD(\Pi^{-1}_i) \defn \cD_{t_i(\Pi^{-1}_i)}(\Pi^{-1}_i)
\end{align}
to emphasize this dependence on the insertion position $\Pi^{-1}_i$.

\item 

Now fix an iteration $s\in[L]$ and condition on $(\cW^{\smallsetminus i}_{s-1},X^{\cW^{\smallsetminus i}_{s-1}})$ and $\Pi_{\smallsetminus i}$.
Under this conditioning, the auxiliary unmasking set $\cD_s^{\smallsetminus i}$ and its size $D^{\smallsetminus i}_s$ are deterministic. 
In the original sampling process, the set of  insertion positions $p$ of token $i$ in the permutation $\Pi$ that lead to $t_i(p) = s$ is precisely the unmasking set $\cD^{\smallsetminus i}_s$ in the auxiliary process, i.e.,
\begin{align*}
\{p\in[L]: t_i(p) = s\} = \cD^{\smallsetminus i}_s.
\end{align*}
To see this, note that each index except the last one in $ \cD^{\smallsetminus i}_s$ has the property that the sum up to the following token is still below the threshold. Therefore, token $i$ can be inserted at any such position without changing the iteration at which it is unmasked. On the other hand, the last index in $\cD^{\smallsetminus i}_s$ is the one at which the cumulative sum including the following token exceeds the threshold. Therefore, for token $i$ to be unmasked at iteration $s$, this is the last position at which it can be inserted.

\item However, if token $i$ is inserted at position $p\in \cD^{\smallsetminus i}_s$ so that it is unmasked at iteration $s$ in the original sampling process, the resulting unmasking set $\cD_s$ may still vary with $p$, because a high entropy of token $i$ can lead to early termination of iteration $s$. Accordingly, for positions~$p\in \cD^{\smallsetminus i}_s$, we write
\begin{align*}
D_s(p) = |\cD_s(p)|
\end{align*}
to denote the size of the original unmasking set at iteration $s$ when token $i$ is inserted at position $p$ in the permutation $\Pi$.

More precisely, the set of all other unmasked tokens in the same iteration $s$ as token $i$ in the original process, is a subset of the $s$-th unmasking set in the auxiliary process, i.e.,
\begin{align}\label{eq:unmasking-set-inclusion}
\cD(p) \smallsetminus \{i\} = \cD_s(p) \smallsetminus \{i\}  \subseteq  \cD^{\smallsetminus i}_s, \quad \forall\,p \in \cD^{\smallsetminus i}_s.
\end{align}
This is because inserting token $i$ into position $p\in \cD^{\smallsetminus i}_s$ in the permutation $\Pi$ can only make iteration~$s$ terminate earlier in the original process than in the auxiliary process since the entropy of token $i$ can be large.

\item
To organize these insertion positions, it is convenient to index them relative to the start of the $s$-th auxiliary set $\cD_s^{\smallsetminus i}$. 
Recall $W_s^{\smallsetminus i} \defn |\cW_s^{\smallsetminus i}|$ and $D_s^{\smallsetminus i} \defn |\cD_s^{\smallsetminus i}|$.
For each absolute insertion position $q\in[L]$, let 
\begin{align*}
r \defn \big((q-1) \bmod D_s^{\smallsetminus i}\big) + 1,
\end{align*}
be the relative position of $q$ within the set $\cD^{\smallsetminus i}_s$, and define
\begin{align}
D^\rel_s(q) \defn D_s(r+W_{s-1}^{\smallsetminus i}). \label{eq:relative-size-function}
\end{align}
In other words, $D_s^{\rel}(q)$ represents the size of the $s$-th original unmasking set $\cD_s$ obtained when token~$i$ is inserted at relative position $r$ within the $s$-th auxiliary unmasking set $\cD^{\smallsetminus i}_s$. In particular, we introduce the module operation in \eqref{eq:relative-size-function} to handle the case when $q$ is larger than the auxiliary set size $D_s^{\smallsetminus i}$.

\item Finally, note that conditioned on $t_i(p) = s$, the envelope  $\ol D_s$ is determined by the original process up to iteration $s-1$, and therefore does not depend on the relative index $q$ of the insertion position $p$ within the $s$-th auxiliary set $\cD_s^{\smallsetminus i}$.

\end{itemize}


\paragraph{Proof of Lemma \ref{lem:entropy-sum-token-i-kl-bound}.}

Having the above preparations, we are ready to prove Lemma \ref{lem:entropy-sum-token-i-kl-bound}.

Our goal is to analyze the contribution of token $i$ to the expected KL sampling error in \eqref{eq:kl-low-dim-tower-sum-i-1} by taking expectation with respect to the randomness of $\Pi^{-1}_i$, namely, the random insertion position $\Pi_i^{-1}$ of token $i$ in the permutation $\Pi$. Hence, for each $p\in[L]$, define
\begin{align}
F(p) &\defn \ptI(X^{i}; X^{\cD(p) \smallsetminus \{i\}} \mid \cW^{\smallsetminus i}_{t_i(p)-1},X^{\cW^{\smallsetminus i}_{t_i(p)-1}},\Pi_{\smallsetminus i}) \ind{\big\{\ptH (X^i \mid \cW^{\smallsetminus i}_{t_i(p)-1},X^{\cW^{\smallsetminus i}_{t_i(p)-1}} ) \leq \eta \big\}}.
\end{align}
We can then express \eqref{eq:kl-low-dim-tower-sum-i-1} as
\begin{align}\label{eq:kl-low-dim-tower-sum-i-1-expression}
& \bE_{X, \Pi}\Big[\ptI(X^{i}; X^{\cD_{t_i}\smallsetminus \{i\}} \mid \cW_{t_i-1},X^{\cW_{t_i-1}},\Pi) \ind{\big\{\ptH(X^i \mid \cW_{t_i-1},X^{\cW_{t_i-1}}) \leq \eta\big\}}\Big] \notag\\
& \qquad = \bE\big[F(\Pi^{-1}_i)\big] = \bE_{X^{\smallsetminus i}, \Pi_{\smallsetminus i}}\Big[ \bE_{\Pi^{-1}_i}\big[ F(\Pi^{-1}_i) \mid X^{\smallsetminus i}, \Pi_{\smallsetminus i} \big] \Big].
\end{align}

In what follows, we focus on the inner expectation with respect to $\Pi_i^{-1}$ in \eqref{eq:kl-low-dim-tower-sum-i-1-expression}.
As $\Pi_i^{-1}$ is uniformly distributed over $[L]$ conditioned on $X^{\smallsetminus i}$ and~$\Pi_{\smallsetminus i}$, we have
\begin{align}
& \bE_{\Pi^{-1}_i}\big[ F(\Pi^{-1}_i) \mid X^{\smallsetminus i}, \Pi_{\smallsetminus i} \big] \notag \\
& \quad \numpf{i}{=} \frac1L \sum_{p=1}^L \ptI(X^{i}; X^{\cD(p) \smallsetminus \{i\}} \mid \cW^{\smallsetminus i}_{t_i(p)-1},X^{\cW^{\smallsetminus i}_{t_i(p)-1}},\Pi_{\smallsetminus i}) \ind{\big\{\ptH (X^i \mid \cW^{\smallsetminus i}_{t_i(p)-1},X^{\cW^{\smallsetminus i}_{t_i(p)-1}} ) \leq \eta \big\}} \notag \\ 
& \quad \numpf{ii}{=} \frac1L \sum_{s=1}^L \sum_{p=1}^L\ind\{t_i(p) = s\} \ptI(X^{i}; X^{\cD(p) \smallsetminus \{i\}} \mid \cW^{\smallsetminus i}_{t_i(p)-1},X^{\cW^{\smallsetminus i}_{t_i(p)-1}},\Pi_{\smallsetminus i}) \ind{\big\{\ptH (X^i \mid \cW^{\smallsetminus i}_{t_i(p)-1},X^{\cW^{\smallsetminus i}_{t_i(p)-1}} ) \leq \eta \big\}} 
\notag \\
& \quad \numpf{iii}{=} \frac1L \sum_{s=1}^L \sum_{p\in  \cD^{\smallsetminus i}_s} \ptI(X^{i}; X^{\cD(p) \smallsetminus \{i\}} \mid \cW_{s-1}^{\smallsetminus i},X^{\cW_{s-1}^{\smallsetminus i}},\Pi_{\smallsetminus i}) \ind{\big\{\ptH (X^i \mid \cW_{s-1}^{\smallsetminus i},X^{\cW_{s-1}^{\smallsetminus i}}) \leq \eta \big\}} \notag\\
& \quad \numpf{iv}{=} \frac1L \sum_{s=1}^L \sum_{q=1}^{D_s^{\smallsetminus i}} \ptI(X^{i}; X^{\cD(q+W_{s-1}^{\smallsetminus i}) \smallsetminus \{i\}} \mid \cW_{s-1}^{\smallsetminus i},X^{\cW_{s-1}^{\smallsetminus i}},\Pi_{\smallsetminus i}) \ind{\big\{\ptH (X^i \mid \cW_{s-1}^{\smallsetminus i},X^{\cW_{s-1}^{\smallsetminus i}}) \leq \eta \big\}}
\label{eq:kl-low-dim-tower-sum-i-2}
\end{align}
Here, (i) follows from the fact that $\Pi^{-1}_i$ is uniformly distributed over $[L]$ independent of $X^{\smallsetminus i}$ and $\Pi_{\smallsetminus i}$ and that conditioned on $X^{\smallsetminus i}$ and $\Pi_{\smallsetminus i}$, $t_i$ is determined given $\Pi^{-1}_i = p$; (ii) rearranges the summation by grouping the insertion positions with the same iteration index $t_i$; (iii) arises from the fact $\{p\in[L]: t_i(p) = s\} = \cD^{\smallsetminus i}_s$; (iv) rewrites the position $p$ in terms of the relative position $q$ within the auxiliary unmasking set $\cD^{\smallsetminus i}_s$.

Next, let us swap the order of double summation in \eqref{eq:kl-low-dim-tower-sum-i-2}.
The technical challenge is that the size of the second summation, or equivalently the size of the auxiliary unmasking set $\cD^{\smallsetminus i}_s$, varies across different iterations $s$. To ensure that the swapped summation is well-organized, we can leverage the dyadic size envelopes $(\ol{D}_s)_{s=1}^L$ to obtain an upper bound.
This is established by the following lemma, whose proof is deferred to Appendix~\ref{sec:proof-sum-swap}.
\begin{lemma}\label{lem:sum-swap}
    Recall the definition of $\ol{D}_s$ and $D^\rel_s(q)$ in \eqref{eq:size-envelope-sequence} and \eqref{eq:relative-size-function}, respectively. The sum in \eqref{eq:kl-low-dim-tower-sum-i-2} satisfies
\begin{align}\label{eq:kl-low-dim-tower-sum-i-3}
&\sum_{s=1}^L \sum_{q=1}^{D_s^{\smallsetminus i}} \ptI(X^{i}; X^{\cD(q+W_{s-1}^{\smallsetminus i}) \smallsetminus \{i\}} \mid \cW_{s-1}^{\smallsetminus i},X^{\cW_{s-1}^{\smallsetminus i}},\Pi_{\smallsetminus i}) \ind{\big\{\ptH (X^i \mid \cW_{s-1}^{\smallsetminus i},X^{\cW_{s-1}^{\smallsetminus i}}) \leq \eta \big\}} \notag \\
&\quad \le \sum_{s=1}^L \sum_{k=s}^L \sum_{q = 1}^{2 D^{\smallsetminus i}_s \wedge  D^{\smallsetminus i}_k}  \ptI(X^{i}; X^{\cD(q+W_{k-1}^{\smallsetminus i}) \smallsetminus \{i\}} \mid \cW_{k-1}^{\smallsetminus i},X^{\cW_{k-1}^{\smallsetminus i}},\Pi_{\smallsetminus i}) \notag\\
& \hspace{10em} \cdot \ind\{ D^\rel_{s}(q) \geq \ol{D}_s\}\cdot\ind{\big\{\ptH(X^i \mid \cW_{k-1}^{\smallsetminus i},X^{\cW_{k-1}^{\smallsetminus i}}) \leq \eta \big\}}. 
\end{align}
\end{lemma}

Equipped with Lemma~\ref{lem:sum-swap}, we can continue the derivation.
Recall \eqref{eq:unmasking-set-inclusion} that
\begin{align*}
\cD(p) \smallsetminus \{i\} \subseteq \cD^{\smallsetminus i}_k, \quad \forall\, p\in \cD^{\smallsetminus i}_k,
\end{align*}
or equivalently, rewriting the absolute position $p$ in terms of the relative position $q$ within the auxiliary unmasking set $\cD^{\smallsetminus i}_k$, we have
\begin{align*}
\cD(q+W_{k-1}^{\smallsetminus i}) \smallsetminus \{i\} \subseteq \cD^{\smallsetminus i}_k, \quad \forall\, 1 \leq q \leq D^{\smallsetminus i}_k.
\end{align*}
Using the monotonicity of mutual information, we can upper bound the right-hand side of \eqref{eq:kl-low-dim-tower-sum-i-3} as
\begin{align}\label{eq:kl-low-dim-tower-sum-i-4}
& \sum_{s=1}^L \sum_{k=s}^L \sum_{q = 1}^{2D^{\smallsetminus i}_s \wedge D^{\smallsetminus i}_k} \ind\{ D^\rel_{s}(q) \geq \ol{D}_s\} \ptI(X^{i}; X^{\cD(q+W_{k-1}^{\smallsetminus i}) \smallsetminus \{i\}} \mid \cW_{k-1}^{\smallsetminus i},X^{\cW_{k-1}^{\smallsetminus i}},\Pi_{\smallsetminus i}) \ind{\big\{\ptH (X^i \mid \cW_{k-1}^{\smallsetminus i},X^{\cW_{k-1}^{\smallsetminus i}} ) \leq \eta \big\}} \notag \\
&\quad \le \sum_{s=1}^L \sum_{k=s}^L \sum_{q = 1}^{2D^{\smallsetminus i}_s \wedge D^{\smallsetminus i}_k} \ind\{ D^\rel_{s}(q) \geq \ol{D}_s\} \ptI(X^{i}; X^{ \cD^{\smallsetminus i}_k} \mid \cW_{k-1}^{\smallsetminus i},X^{\cW_{k-1}^{\smallsetminus i}},\Pi_{\smallsetminus i}) \ind{\big\{\ptH (X^i \mid \cW_{k-1}^{\smallsetminus i},X^{\cW_{k-1}^{\smallsetminus i}} ) \leq \eta \big\}} \notag \\
&\quad \leq \sum_{s=1}^L \sum_{q = 1}^{2D^{\smallsetminus i}_s} \ind\{ D^\rel_{s}(q) \geq \ol{D}_s\} \sum_{k=s}^L  \ptI(X^{i}; X^{ \cD^{\smallsetminus i}_k} \mid \cW_{k-1}^{\smallsetminus i},X^{\cW_{k-1}^{\smallsetminus i}},\Pi_{\smallsetminus i}) \ind{\big\{\ptH (X^i \mid \cW_{k-1}^{\smallsetminus i},X^{\cW_{k-1}^{\smallsetminus i}}) \leq \eta \big\}} \notag\\
&\quad = 2 \sum_{s=1}^L \sum_{q = 1}^{D^{\smallsetminus i}_s} \ind\{ D^\rel_{s}(q) \geq \ol{D}_s\} \sum_{k=s}^L  \ptI(X^{i}; X^{ \cD^{\smallsetminus i}_k} \mid \cW_{k-1}^{\smallsetminus i},X^{\cW_{k-1}^{\smallsetminus i}},\Pi_{\smallsetminus i}) \ind{\big\{\ptH (X^i \mid \cW_{k-1}^{\smallsetminus i},X^{\cW_{k-1}^{\smallsetminus i}}) \leq \eta \big\}} \notag\\
& \quad = 2 \sum_{s=1}^L  \sum_{p=1}^L \ind\{t_i(p) = s\} \ind\{ D_s(p) \geq \ol{D}_s\} \sum_{k=s}^L \ptI(X^{i}; X^{ \cD^{\smallsetminus i}_k} \mid \cW_{k-1}^{\smallsetminus i},X^{\cW_{k-1}^{\smallsetminus i}},\Pi_{\smallsetminus i}) \ind{\big\{\ptH (X^i \mid \cW_{k-1}^{\smallsetminus i},X^{\cW_{k-1}^{\smallsetminus i}}) \leq \eta \big\}},
\end{align}
where the penultimate step is due to the definition that $D^\rel_s(q)$ is a function of $q$ modulo $D^{\smallsetminus i}_s$, and the last step is true because for $q \in [D^{\smallsetminus i}_s]$, it corresponds to an insertion position $p$ such that $t_i(p) = s$ with $D_s(p)=D^\rel_s(q)$.

Putting \eqref{eq:kl-low-dim-tower-sum-i-3}--\eqref{eq:kl-low-dim-tower-sum-i-4} together and substituting it into \eqref{eq:kl-low-dim-tower-sum-i-2}, we obtain that conditioned on $X^{\smallsetminus i}$ and $\Pi_{\smallsetminus i}$, 
\begin{align}
& \bE_{\Pi^{-1}_i}\big[ F(\Pi^{-1}_i) \mid X^{\smallsetminus i}, \Pi_{\smallsetminus i} \big] \notag \\
&\qquad \leq \frac2L \sum_{s=1}^L  \sum_{p=1}^L \ind\{t_i(p) = s\} \ind\{D_s(p) \geq \ol{D}_s\} \notag\\
& \hspace{8em} \cdot \sum_{k=s}^L \ptI(X^{i}; X^{ \cD^{\smallsetminus i}_k} \mid \cW_{k-1}^{\smallsetminus i},X^{\cW_{k-1}^{\smallsetminus i}},\Pi_{\smallsetminus i}) \ind{\big\{\ptH \big(X^i \mid \cW_{k-1}^{\smallsetminus i},X^{\cW_{k-1}^{\smallsetminus i}},\Pi_{\smallsetminus i} \big) \leq \eta \big\}}.\label{eq:kl-low-dim-tower-sum-i-5}
\end{align}

To control the sum of the mutual information in the above display, the key insight is that the auxiliary unmasking sets satisfy $\cW^{\smallsetminus i}_{k} = \cW^{\smallsetminus i}_{k-1} \cup \cD^{\smallsetminus i}_{k}$, which allows us to leverage the chain rule of mutual information to control the sum. This is accomplished in the following lemma, whose proof can be found in Appendix~\ref{sec:proof-mutual-info-chain-rule-max}.
\begin{lemma}
\label{lem:mutual-info-chain-rule-max}
For any token index $i\in[L]$ and iteration index $s\in[L]$, one has
\begin{align}\label{eq:mi-chain-rule-2}
    & \bE\Bigg[\sum_{k=s}^L \ptI(X^{i}; X^{ \cD^{\smallsetminus i}_k} \mid \cW_{k-1}^{\smallsetminus i},X^{\cW_{k-1}^{\smallsetminus i}},\Pi_{\smallsetminus i}) \ind{\big\{\ptH(X^i \mid \cW_{k-1}^{\smallsetminus i},X^{\cW_{k-1}^{\smallsetminus i}}) \leq \eta \big\}} \,\big|\,\cW_{s-1}^{\smallsetminus i},X^{\cW_{s-1}^{\smallsetminus i}},\Pi_{\smallsetminus i} \Bigg] \notag \\
    & \qquad \leq \ptH (X^i \mid \cW_{s-1}^{\smallsetminus i},X^{\cW_{s-1}^{\smallsetminus i}}) \wedge \eta.
\end{align}
\end{lemma}

With the help of Lemma~\ref{lem:mutual-info-chain-rule-max}, let us compute the expectation over $X^{\smallsetminus i}$ and $\Pi_{\smallsetminus i}$ on both sides of \eqref{eq:kl-low-dim-tower-sum-i-5}.
For each $s\in[L]$, the expectation of the $s$-th term on the right-hand-side of \eqref{eq:kl-low-dim-tower-sum-i-5} can be upper bounded as
\begin{align}
& \bE\Bigg[\bE\bigg[\sum_{p=1}^L \ind\{t_i(p) = s\} \ind\{D_s(p) \geq \ol{D}_s\} \notag\\
& \quad\qquad \cdot \sum_{k=s}^L \ptI(X^{i}; X^{ \cD^{\smallsetminus i}_k} \mid \cW_{k-1}^{\smallsetminus i},X^{\cW_{k-1}^{\smallsetminus i}},\Pi_{\smallsetminus i}) \ind{\big\{\ptH \big(X^i \mid \cW_{k-1}^{\smallsetminus i},X^{\cW_{k-1}^{\smallsetminus i}} \big) \leq \eta \big\}} \,\Big|\, \cW_{s-1}^{\smallsetminus i},X^{\cW_{s-1}^{\smallsetminus i}},\Pi_{\smallsetminus i} \bigg] \Bigg] \notag\\
& \quad \numpf{i}{\leq} \bE\Bigg[  \sum_{p=1}^L \ind\{t_i(p) = s\}\ind\{D_s(p) \geq \ol{D}_s\} \notag\\
& \qquad \qquad \cdot \bE\bigg[ \sum_{k=s}^L \ptI(X^{i}; X^{ \cD^{\smallsetminus i}_k} \mid \cW_{k-1}^{\smallsetminus i},X^{\cW_{k-1}^{\smallsetminus i}},\Pi_{\smallsetminus i}) \ind\big\{\ptH (X^i \mid \cW_{k-1}^{\smallsetminus i},X^{\cW_{k-1}^{\smallsetminus i}}) \leq \eta \big\} \mid \cW_{s-1}^{\smallsetminus i},X^{\cW_{s-1}^{\smallsetminus i}},\Pi_{\smallsetminus i} \bigg] \Bigg]\notag\\
& \quad \numpf{ii}{\leq}  \bE\Bigg[  \sum_{p=1}^L \ind\{t_i(p) = s\} \ind\big\{D_s(p) \geq \ol{D}_s\big\} \big( \ptH (X^i \mid \cW_{s-1}^{\smallsetminus i},X^{\cW_{s-1}^{\smallsetminus i}}) \wedge \eta \big) \Bigg], \notag
\end{align}
where (i) holds because $\{t_i(p)=s\}$ and $\{D_s(p) \geq \ol{D}_s\}$ are determined by $(\cW_{s-1}^{\smallsetminus i},X^{\cW_{s-1}^{\smallsetminus i}})$ and $\Pi_{\smallsetminus i}$; (ii) applies \eqref{eq:mi-chain-rule-2} in Lemma~\ref{lem:mutual-info-chain-rule-max}.
Summing the above bound over $s\in[L]$ yields
\begin{align*}
\bE\big[ F(\Pi^{-1}_i) \big]
& \leq  \frac2L \bE\Bigg[ \sum_{s=1}^L  \sum_{p=1}^L \ind\{t_i(p) = s\} \ind\big\{D_s(p) \geq \ol{D}_s\big\} \big( \ptH (X^i \mid \cW_{s-1}^{\smallsetminus i},X^{\cW_{s-1}^{\smallsetminus i}}) \wedge \eta \big) \Bigg] \notag \\
& \numpf{i}{=} \frac2L \bE\Bigg[ \sum_{p=1}^L \ind\big\{D_{t_i(p)} \geq \ol{D}_{t_i(p)}\big\} \big( \ptH (X^i \mid \cW_{t_i(p)-1},X^{\cW_{t_i(p)-1}}) \wedge \eta \big) \Bigg] \notag \\
& \numpf{ii}{=} 2\,\bE_{X^{\smallsetminus i}, \Pi_{\smallsetminus i}}\Big[\bE_{\Pi_i^{-1}} \big[ \ind\{D_{t_i} \geq \ol{D}_{t_i}\} \big( \ptH(X^i \mid \cW_{t_i-1},X^{\cW_{t_i-1}}) \wedge \eta \big) \mid X^{\smallsetminus i}, \Pi_{\smallsetminus i} \big]\Big], \notag \\
& = 2\,\bE\Big[ \ind\{D_{t_i} \geq \ol{D}_{t_i}\} \big( \ptH(X^i \mid \cW_{t_i-1},X^{\cW_{t_i-1}}) \wedge \eta \big) \Big],
\end{align*}
where (i) swaps the summation order and uses the fact that $\cW^{\smallsetminus i}_{t_i-1} = \cW_{t_i-1}$; (ii) is true because $\Pi_i^{-1}$ is uniformly random over $[L]$ and independent of $X^{\smallsetminus i}$ and $\Pi_{\smallsetminus i}$.

Combining the above bound with \eqref{eq:kl-low-dim-tower-sum-i-1-expression}, we arrive at the desired claim in \eqref{eq:entropy-sum-token-i-kl-bound}.
This completes the proof of Lemma~\ref{lem:entropy-sum-token-i-kl-bound}.

\subsection{Proof of Lemma~\ref{lem:entropy-expression}}
\label{sec:lem:entropy-expression}
We begin with the proof of \eqref{eq:entropy-expression-joint}.
Recall from \eqref{eq:data-density-low-dim} that for a fixed permutation $\Pi=\pi$, the data density can be expressed as
\begin{align*}
\log p_{X_0}(x) =  \sum_{t=1}^{\tau} \log p(x^{d_t}\mid x^{w_{t-1}}),
\end{align*}
where the realized number of iterations $\tau$ and unmasking sets $(d_t, w_{t-1})_{t\in[\tau]}$ are determined by $x=(x^1, \dots, x^L)$ and $\pi$.
One can then apply the same argument as in \eqref{eq:kl-low-dim-tower}. Specifically, taking expectation over $X\sim p_{X_0}$ yields that for any fixed $\Pi=\pi$,
\begin{align*}
H(X_0) & = \bE_{X\sim p_{X_0}}\bigl[-\log p_{X_0}(X)\bigr]
= \bE_{X}\Biggl[-\sum_{t=1}^{T} \log p(X^{d_t}\mid X^{w_{t-1}}) \,\big|\,\Pi=\pi \Biggr] \\
& = \bE_X\Biggl[\sum_{t=1}^{T} \ptH(X^{d_t} \mid w_{t-1}, X^{w_{t-1}}, \pi)\Biggr].
\end{align*} 

Next, let us turn to \eqref{eq:entropy-expression-margin}. Recall the density of the sampled distribution in \eqref{eq:sample-density-low-dim}:
\begin{align*}
\log p_{Y\mid \Pi = \pi}(x) =  \sum_{t=1}^{\tau} \sum_{i\in d_t} \log p(x^{i}\mid x^{w_{t-1}}).
\end{align*}
We can then express the entropy of $X_0$ as
\begin{align*}
H(X_0) & = \bE_{X\sim p_{X_0}}\biggl[- \log p_{Y_T\mid \Pi = \pi}(X) -\log \frac{p_{X_0}(X)}{p_{Y_T\mid \Pi = \pi}(X)} \biggr] \notag\\
& = \bE_{X}\bigl[-\log p_{Y_T\mid \Pi = \pi}(X)\bigr] - \KL(p_{X_0} \parallel p_{Y_T\mid \Pi = \pi}) \\
& = \bE_X\Biggl[\sum_{t=1}^{T} \sum_{i\in \cD_t} \ptH(X^{i} \mid \cW_{t-1}, X^{\cW_{t-1}}, \pi)\Biggr] - \KL(p_{X_0} \parallel p_{Y_T\mid \Pi = \pi}).
\end{align*}

\subsection{Proof of Lemma \ref{lem:sum-swap}}
\label{sec:proof-sum-swap}
We prove Lemma \ref{lem:sum-swap} by showing that every term on the left-hand-side of \eqref{eq:kl-low-dim-tower-sum-i-3} is covered by a term on the right-hand-side.

Fix an arbitrary $1 \leq k \leq L$ and $1 \leq q \leq D^{\smallsetminus i}_k$. 
Recall the definitions of $\ol{D}_s$ and $D_s^\rel(q)$ in \eqref{eq:size-envelope-sequence} and \eqref{eq:relative-size-function}, respectively. Define
\begin{align}\label{eq:def-s}
s = s(k,q) \defn \max \big\{u\leq k \,:\, \ol{D}_u \leq q\big\}.
\end{align}
This set is non-empty because $\ol{D}_1 = 0 $, so $s$ is well-defined.

In what follows, let us show that the right-hand-side term with the index $(s,k,q)$ matches the term indexed by $(k,q)$ on the left-hand-side.

To begin with, let us verify that the indicator $\ind\{ D^\rel_{s}(q) \geq \ol{D}_s\}$ is $1$.
We first claim that the relative size $D_s^\rel(q)$ is at least $q$, i.e.,
\begin{align}\label{eq:relative-size-lower-bound}
D_s^\rel(q) \geq q.
\end{align}
To see this, recall that any position $1 \leq q \leq D^{\smallsetminus i}_s$ guarantees that token $i$ is unmasked at iteration $s$. As the sampling process must first unmask the first $q-1$ tokens in the set and then unmask token $i$, the unmasking set size must satisfy $D_s^\rel(q) = |\cD_s(q)| \geq q$. This establishes \eqref{eq:relative-size-lower-bound}.

We can then combine \eqref{eq:relative-size-lower-bound} with the definition of $s$ in \eqref{eq:def-s} to find
\begin{align*}
D_s^\rel(q) \geq q \geq \ol{D}_s.
\end{align*}
Thus the indicator $\ind\{ D^\rel_{s}(q) \geq \ol{D}_s\}$ is $1$.

As we choose $q\leq D_k^{\smallsetminus i}$, it remains to prove that 
\begin{align*}
q \leq 2 D^{\smallsetminus i}_s.
\end{align*}

If $s = k$, then the above inequality holds trivially because $q \leq D^{\smallsetminus i}_k \leq 2 D^{\smallsetminus i}_s$.

If $s < k$, by the definition of $s$ in \eqref{eq:def-s}, we must have $\ol{D}_{s+1} > q \geq \ol{D}_s$. On the other hand, by the recursive definition of $\ol{D}_{s+1}$ in \eqref{eq:size-envelope-sequence}, $\ol{D}_{s+1}$ can only take a value in $\{\ol{D}_{s}, 2D_s\}$. This implies that $\ol{D}_{s+1} = 2 D_s$. Moreover, because $q \leq D_k^{\smallsetminus i}$ guarantees that token $i$ is unmasked at iteration $k$ with $k > s$, we know that $D_s = D^{\smallsetminus i}_s$ by by Observation~\ref{obs:coincide}. Therefore, this leads to
$$\ol{D}_{s+1} = 2 D_s = 2 D^{\smallsetminus i}_s > q.$$

Hence, in either case, we have $$q \leq 2 D^{\smallsetminus i}_s \wedge D^{\smallsetminus i}_k. $$ 

This proves that each term indexed by $(k,q)$ on the left-hand-side corresponds to a term indexed by $(s,k,q)$ on the right-hand-side, and the proof of Lemma~\ref{lem:sum-swap} is complete.

\subsection{Proof of Lemma~\ref{lem:mutual-info-chain-rule-max}}
\label{sec:proof-mutual-info-chain-rule-max}

Fix a token index $i$ and an iteration index $s$. 
Before proceeding, we first summarize some notation and observations.
For convenience of notation, for each $s \leq k \leq L$, define the filtration
\[\mathcal{F}_{k-1} := \sigma\bigl(\cW_{k-1}^{\smallsetminus i},X^{\cW_{k-1}^{\smallsetminus i}},\Pi_{\smallsetminus i}\bigr).
\]
In addition, denote the pointwise conditional entropy and mutual information by
\begin{align*}
 h_{k-1} & := \ptH(X^i \mid \mathcal{F}_{k-1})  = \ptH(X^i \mid \cW_{k-1}^{\smallsetminus i},X^{\cW_{k-1}^{\smallsetminus i}},\Pi_{\smallsetminus i}) = \ptH(X^i \mid \cW_{k-1}^{\smallsetminus i},X^{\cW_{k-1}^{\smallsetminus i}}), \\
\Delta_k &:= \ptI(X^i; X^{\mathcal D_k^{\smallsetminus i}} \mid \mathcal{F}_{k-1}) = \ptI(X^i; X^{\mathcal D_k^{\smallsetminus i}} \mid \cW_{k-1}^{\smallsetminus i},X^{\cW_{k-1}^{\smallsetminus i}},\Pi_{\smallsetminus i}).
\end{align*}

By the chain rule for mutual information, we can express the conditional mutual information $\Delta_k$ as
\begin{align}
    \label{eq:mi-chain-rule-2-telescope}
\Delta_k
& = \ptH(X^i \mid \mathcal{F}_{k-1})
- \bE_{X^{\mathcal D_k^{\smallsetminus i}}}\Big[ \ptH\big( X^i \mid X^{\mathcal D_k^{\smallsetminus i}}, \mathcal{F}_{k-1} \big) \mid \mathcal{F}_{k-1} \Big] \notag \\ 
& = h_{k-1} - \bE\big[ h_k \mid \mathcal{F}_{k-1} \big],
\end{align}
where the second equality follows from the fact that $\mathcal{F}_k = \sigma(\cW_{k}^{\smallsetminus i},X^{\cW_{k}^{\smallsetminus i}},\Pi_{\smallsetminus i}) = \sigma(\mathcal{F}_{k-1}, X^{\mathcal D_k^{\smallsetminus i}})$ because given $\cW_{k}^{\smallsetminus i}$, $X^{\cW_{k}^{\smallsetminus i}}$, and $\Pi_{\smallsetminus i}$, the unmasking set $\mathcal D_k^{\smallsetminus i}$ is determined.
In particular, $\Delta_k \geq 0$ due to the non-negativity of mutual information, and $(h_k)_{k}$ is a nonnegative supermartingale.

Finally, define the first entrance time into the low-entropy region by
\[\tau := \min\{s\leq k \leq L: h_{k-1}\leq \eta\},\]
with the convention that $\tau=\infty$ if the set is empty.

With the above definitions, the identity in \eqref{eq:mi-chain-rule-2} can be rewritten as
\begin{align}
    \label{eq:mi-chain-rule-2-rewrite}
\bE\Bigg[ \sum_{k=s}^{L} \Delta_k \ind\{h_{k-1}\leq\eta\} \mid \cF_{s-1} \Bigg] \leq h_{s-1} \wedge \eta,
\end{align}
and the remainder of the proof focuses on establishing this inequality.

By the definition of $\tau$, every term selected by the indicator $\ind\{h_{k-1}\leq\eta\}$ occurs at or after $\tau$. Combining this with the fact that $\Delta_k\geq 0$, we have
\begin{align}
\label{eq:indicator-first-hitting}
\mathbb{E}\Biggl[\sum_{k=s}^{L}\Delta_k\ind\{h_{k-1}\leq\eta\}\mathrel{\big|}\mathcal{F}_{s-1}\Biggr]\leq\mathbb{E}\Biggl[\ind\{\tau\leq L\}\sum_{k=\tau}^{L}\Delta_k\mathrel{\big|}\mathcal{F}_{s-1}\Biggr].
\end{align}
Since $h_{k-1}=\ptH(X^i \mid \mathcal{F}_{k-1})$ is $\mathcal{F}_{k-1}$-measurable, we have
$\{\tau=k\}\in\mathcal{F}_{k-1}$ for every $s\le k\le L$. Decomposing the right-hand side of \eqref{eq:indicator-first-hitting} over the value of $\tau$ and using the tower property with $\cF_{s-1}\subseteq\cF_{k-1}$ yields
\begin{align*}
\bE\Biggl[\ind\{\tau\le L\}\sum_{k=\tau}^{L}\Delta_k \Bigm| \cF_{s-1}\Biggr]
&=\sum_{k=s}^{L}\bE\Biggl[\ind\{\tau=k\}\,
   \bE\biggl[\sum_{j=k}^{L}\Delta_j \Bigm| \cF_{k-1}\biggr] \Bigm| \cF_{s-1}\Biggr]\\
&=\sum_{k=s}^{L}\bE\Bigl[\ind\{\tau=k\}
   \bigl(h_{k-1}-\bE[h_L\mid\cF_{k-1}]\bigr) \bigm| \cF_{s-1}\Bigr]\\
&\le\sum_{k=s}^{L}\bE\bigl[\ind\{\tau=k\}\,h_{k-1} \bigm| \cF_{s-1}\bigr] \\
 & \le\eta\,\bP\{\tau\le L\mid\cF_{s-1}\}\le\eta,
\end{align*}
where the second equality uses the telescoping identity \eqref{eq:mi-chain-rule-2-telescope} from $k$ to $L$, the first inequality uses $h_L\ge0$, and the second inequality holds because $h_{k-1}\le\eta$ when $\tau=k$.

On the other hand, dropping the indicator and applying the same telescoping identity from iteration $s$ gives
\[\mathbb{E}\Biggl[\sum_{k=s}^{L}\Delta_k\ind\{h_{k-1}\leq\eta\}\mathrel{\big|}\mathcal{F}_{s-1}\Biggr]\leq\mathbb{E}\Biggl[\sum_{k=s}^{L}\Delta_k\mathrel{\big|}\mathcal{F}_{s-1}\Biggr]=h_{s-1}-\mathbb{E}\bigl[h_L\mid\mathcal{F}_{s-1}\bigr]\leq h_{s-1}.\]
Combining the two bounds, we obtain \eqref{eq:mi-chain-rule-2-rewrite}, which completes the proof of Lemma~\ref{lem:mutual-info-chain-rule-max}.
\section{Proof of Proposition~\ref{prop:hard-instance-sorted-ordering}}
\label{sec:hard-instance-sorted-ordering}
Let $m=\lfloor H/\eta\rfloor$ and define
\[
\delta_k
=
\frac{\delta\eta}{\log(k+1)},
\qquad
1\leq k\leq m,
\]
where $\delta>0$ is sufficiently small. For each $1\leq k\leq m$, let $X_0^{2k-1}=X_0^{2k}=Z_k,$
where the random variables $Z_1,\ldots,Z_m$ are independent and $Z_k\sim\operatorname{Bern}(p_k),$
with $p_k$ chosen so that
\[
-p_k\log p_k-(1-p_k)\log(1-p_k)
=
\eta-\delta_k.
\]
All remaining coordinates are deterministic: $X_0^j=0,
j>2m.$

For $1\leq k\leq m$, we have
\[
H(X_0^{2k-1})
=
H(X_0^{2k})
=
H(X_0^{2k-1},X_0^{2k})
=
\eta-\delta_k,
\]
and hence
\[
I(X_0^{2k-1};X_0^{2k})
=
\eta-\delta_k.
\]
Therefore,
\[
H(X_0)
=
\sum_{k=1}^m H(Z_k)
=
\sum_{k=1}^m(\eta-\delta_k)
<
m\eta
\leq H.
\]

Under sorted ordering, all deterministic coordinates, whose conditional entropy is zero, are selected first. The first correlated pair is then added to the same iteration. Since
\[
H(X_0^{2k-1})=\eta-\delta_k<\eta
\]
but
\[
H(X_0^{2k-1})+H(X_0^{2k})
=
2(\eta-\delta_k)>\eta
\]
for sufficiently small $\delta$, each subsequent iteration contains exactly one correlated pair:
\[
D_1
=
\{j>2m\}\cup\{1,2\},
\qquad
D_t
=
\{2t-1,2t\},
\quad
2\leq t\leq m.
\]
Consequently, the number of iterations is
\[
T=m
=
\biggl\lfloor\frac{H}{\eta}\biggr\rfloor.
\]

Moreover, each pair is decoded independently within the same iteration, so its contribution to the KL error is its mutual information. Hence,
\[
\KL(p_{X_0}\|p_Y)=
\sum_{t=1}^m
I(X_0^{2t-1};X_0^{2t})=
\sum_{t=1}^m(\eta-\delta_t)=
\Theta(H).
\]
This construction shows that selecting tokens solely according to low marginal entropy does not guarantee small sampling error.

\section{Proof of Theorem~\ref{thm:maximum}}
\label{sec:max-entropy-analysis}

\subsection{Proof sketch of Theorem~\ref{thm:maximum}}
The high-level proof strategy for Theorem~\ref{thm:maximum} is similar to that for Theorem~\ref{thm:entropy-sum-sampling}. Hence, we will mainly focus on the key differences in the analysis, and refer readers to Section~\ref{sec:entropy-sum-analysis} for the common technical details.

\paragraph{Step 1: Decompose KL error into sum of mutual information.}

To begin with, notice that the sampling process of the maximum entropy-based strategy in Algorithm~\ref{alg:entropy-based-sampling-max} also admits a Markovian structure:
\begin{align}
& \Pi \to \bigl\{\ptH\big(p^{\Pi_k}_\theta(\cdot \mid Y_{0}) \big)\bigr\}_{k=1}^L \to (\cD_1,\cW_1) \to Y_1 \to \bigl\{\ptH\big(p^{\Pi_k}_\theta(\cdot \mid Y_1) \big)\bigr\}_{k= W_1+1}^L \to (\cD_2,\cW_2) \to Y_2  \notag \\
&\qquad \to  \dots
\to \bigl\{\ptH\big(p^{\Pi_k}_\theta(\cdot \mid Y_{T-1}) \big)\bigr\}_{k=W_{T-1}+1}^L \to (\cD_{T},\cW_{T}) \to Y_T. \label{eq:markov-chain-entropy-max}
\end{align}

Hence, the KL sampling error takes the same form as \eqref{eq:kl-low-dim-tower-sum-expression} in Section~\ref{sec:entropy-sum-analysis}, i.e.,
\begin{align*}
\bE_{\Pi}\big[\KL(p_{X_0} \parallel p_{Y_{T}\mid \Pi})\big] 
& = \bE_{\Pi,X\sim p_{X_0}}\Bigg[ \sum_{t=1}^T \KL \Big(p^{\cD_t}( \cdot \mid X^{\cW_{t-1}}) \,\big\|\, \prod_{i \in \cD_t} p^i(\cdot \mid  X^{\cW_{t-1}}) \Big) \Bigg].
\end{align*}
It thus suffices to control the per-iteration KL error term in \eqref{eq:kl-low-dim-inner-1} for the maximum entropy-based strategy.

Recall that for the entropy sum-based strategy (Algorithm~\ref{alg:entropy-based-sampling-sum}), we handle this term via the KL divergence bound in \eqref{eq:kl-mutual-info-decomp-2} from Lemma~\ref{lem:kl-mutual-info-decomp}.
Here, we instead apply the KL divergence identity \eqref{eq:kl-mutual-info-decomp} from the same lemma.
Specifically, combining \eqref{eq:kl-mutual-info-decomp} with the representation of the unmasking set $\cD_t$ in \eqref{eq:unmasking-set-representation}:
\begin{align*}
\cD_t = \cW_t \setminus \cW_{t-1} = \{\Pi_{|\cW_{t-1}|+1}, \ldots, \Pi_{|\cW_t|}\},
\end{align*}
we can express the KL error from iteration $t$ in \eqref{eq:kl-low-dim-inner-1} as
\begin{align}
& \KL \Big(p^{d_t}( \cdot \mid x^{w_{t-1}}) \,\big\|\, \prod_{i \in d_t} p^i(\cdot \mid  x^{w_{t-1}}) \Big) \notag\\
 & \qquad = \sum_{k=1}^{|d_t|-1} I\big(X^{\pi_{|w_{t-1}|+k}}; X^{\pi_{|w_{t-1}|+k+1}}, \ldots, X^{\pi_{|w_{t}|}} \,\big|\, (\cW_{t-1},X^{\cW_{t-1}}) = (w_{t-1},x^{w_{t-1}}),\Pi=\pi\big). \label{eq:kl-low-dim-inner}
\end{align}
That is, the contribution of iteration $t$ is given by a sum of mutual information terms, where the $k$-th term measures the mutual information between the $k$-th unmasked token and the tokens that are unmasked later within the same iteration. 
In particular, the last unmasked token in the iteration $X^{\pi_{|w_{t}|}}$ does not contribute to the KL error.

Substituting \eqref{eq:kl-low-dim-inner} back into the expression of the KL sampling error \eqref{eq:kl-low-dim-tower-sum-expression} yields
\begin{align}
& \KL(p_{X_0} \parallel p_{Y_{T}\mid \Pi}) \notag\\
& \quad = \bE \Bigg[ \sum_{t=1}^T \sum_{k=1}^{|\cD_t|-1} \ptI (X^{\Pi_{|\cW_{t-1}|+k}}; X^{\Pi_{|\cW_{t-1}|+k+1}},\dots,X^{\Pi_{|\cW_{t}|}} \mid \cW_{t-1},X^{\cW_{t-1}},\Pi ) \Bigg] \notag\\
& \quad = \bE\Bigg[ \sum_{k=1}^L \ptI (X^{\Pi_k};X^{\Pi_{k+1}},\dots,X^{\Pi_{|\cW_{t_{\Pi_k}}|}} \mid \cW_{t_{\Pi_k}-1}, X^{\cW_{t_{\Pi_k}-1}},\Pi ) \ind\big\{\ptH (X^{\Pi_k}\mid \cW_{t_{\Pi_k}-1},X^{\cW_{t_{\Pi_k}-1}}) < \eta\big\} \Bigg], \label{eq:kl-low-dim-tower-sum-2}
\end{align}
where the second line reindexes by the absolute unmasking order $k\in[L]$ instead of the relative order within each iteration. Here, we recall that $t_i$ denotes the iteration at which token $i$ is unmasked and thus $t_{\Pi_k}$ stands for the iteration at which the $k$-th unmasked token $\Pi_k$ is revealed. The indicator function arises because only non iteration-ending tokens contribute to the KL error and they satisfy the low-entropy condition by the maximum entropy-based stopping rule.

We next regroup the sum by token index $i\in[L]$ instead of unmasking order $k\in[L]$. To this end, let $\cL^i$ be the index set of the tokens that are unmasked in the same iteration as token $i$ but are revealed after it, namely,
\begin{align}\label{eq:li-definition}
\cL^i \defn \big\{\Pi_{\Pi^{-1}_i+1},\Pi_{\Pi^{-1}_i+2},\dots,\Pi_{|\cW_{t_i}|} \big\}.
\end{align}
Note that $\cL^i$ is a deterministic function of the permutation $\Pi$ and the previous iterate $(\cW_{t_i-1},X^{\cW_{t_i-1}})$.
Rewriting the sum in \eqref{eq:kl-low-dim-tower-sum-2} accordingly yields
\begin{align}\label{eq:kl-low-dim-decomp}
\bE_{\Pi}\big[\KL(p_{X_0} \parallel p_{Y_{T}\mid \Pi})\big] 
& = \sum_{i=1}^L \bE\Big[ \ptI(X^{i};X^{\cL^i} \mid \cW_{t_i-1},X^{\cW_{t_i-1}},\Pi) \ind\big\{\ptH(X^{i}\mid \cW_{t_i-1},X^{\cW_{t_i-1}},\Pi ) < \eta\big\} \Big].
\end{align}

In a nutshell, this demonstrates that the expected KL sampling error can be decomposed into a sum of mutual information terms over token index $i\in[L]$, where each term measures the correlation between token~$i$ and the tokens that are unmasked later within the same iteration, conditioned on the tokens revealed in prior iterations and the permutation.

We note that in contrast to the analogous expression \eqref{eq:kl-low-dim-tower-sum} for the entropy sum-based scheme, the mutual information term here involves only the tokens unmasked after token $i$ within the iteration, i.e., $X^{\cL^i}$, rather than all other tokens in the same set, i.e., $X^{\cD_{t_1}\smallsetminus \{i\}}$.

\paragraph{Step 2: Bound KL error by entropy-based stopping rule.}

Equipped with the KL error decomposition in \eqref{eq:kl-low-dim-decomp}, it suffices to fix a token index $i\in[L]$ and focus on the $i$-th term:
\begin{align}\label{eq:low-dim-mutual-info-target}
\bE_{\Pi,X}\Big[\ptI(X^{i};X^{\cL^i} \mid \cW_{t_i-1},X^{\cW_{t_i-1}},\Pi) \ind\big\{\ptH(X^{i}\mid \cW_{t_i-1},X^{\cW_{t_i-1}},\Pi) < \eta\big\} \Big].
\end{align}

This is established in the following lemma; with proof deferred to Appendix \ref{sec:proof-mutual-info-bound-auxiliary-max}.
\begin{lemma}
\label{lem:mutual-info-bound-auxiliary-max}
For any token index $i\in[L]$, we have
\begin{align}\label{eq:mutual-info-bound-auxiliary-max}
& \bE_{X,\Pi_{\smallsetminus i},\Pi^{-1}_i}\Big[\ptI (X^{i}; X^{\cL^i} \mid \cW_{t_i-1} , X^{\cW_{t_i-1}}, \Pi  )  \ind\big\{\ptH(X^{i} \mid \cW_{t_i-1} , X^{\cW_{t_i-1}}) < \eta\big\} \Big] 
\leq \frac{S_{\max}}{L} \eta.
\end{align}
\end{lemma}

%

With Lemma~\ref{lem:mutual-info-bound-auxiliary-max} in hand, plugging \eqref{eq:mutual-info-bound-auxiliary-max} into \eqref{eq:low-dim-mutual-info-target} and summing over all token indices $i\in[L]$, we find that the expected KL sampling error in \eqref{eq:kl-low-dim-decomp} can be bounded as
\begin{align*}
\bE_{\Pi}\big[\KL(p_{X_0} \parallel p_{Y_{T}\mid \Pi})\big] \leq S_{\max} \eta.
\end{align*}
Setting the parameters as
\begin{align*}
S_{\max} = \sqrt{\frac{\veps L}{H(X_0)}} \qquad \text{and} \qquad \eta = \frac{H(X_0) S_{\max}}{L} = \sqrt{\frac{\veps H(X_0)}{L}}
\end{align*}
yields the desired KL error bound $\bE_{\Pi}\big[\KL(p_{X_0} \parallel p_{Y_{T}\mid \Pi})\big] \leq \veps$.

\paragraph{Step 3: Control expected number of iterations.}

As for the expected total number of iterations $\bE[T]$, it is controlled by the following lemma; its proof is deferred to Appendix~\ref{sec:proof-tau-bound}.
\begin{lemma}\label{lem:tau-bound}
The expected number of iterations $\bE[T]$ is upper bounded as
\begin{align}
\bE[T] \leq \frac{2L}{S_{\max}}+1.
\end{align}
\end{lemma}
In other words, the expected number of iterations is of the same order as if each iteration unmasks exactly $S_{\max}$ tokens.

The claim in \eqref{eq:iteration-complexity-maximum} then follows by plugging in the choice of $S_{\max}$ into the above bound.
We now complete the proof of Theorem~\ref{thm:maximum}.


\subsection{Proof of Lemma~\ref{lem:mutual-info-bound-auxiliary-max}}
\label{sec:proof-mutual-info-bound-auxiliary-max}

Our strategy is first to take the expectation over $\Pi_i^{-1}$, the insertion position of token $i$ in the permutation~$\Pi$.
As in Lemma~\ref{lem:entropy-sum-token-i-kl-bound}, the main technical challenge lies in the statistical dependence between $\cL^i$ (the set of tokens unmasked after token $i$ in iteration $t$) and $\Pi^{-1}_i$, making it difficult to take the expectation over the randomness of~$\Pi$ in \eqref{eq:low-dim-mutual-info-target}.

To this end, similar to the proof for Lemma~\ref{lem:entropy-sum-token-i-kl-bound} in Appendix~\ref{sec:proof-entropy-sum-token-i-kl-bound}, we consider the auxiliary sampling process in which we unmask the entire token sequence except for the token with index~$i$. We work with $X^{\smallsetminus i} = (X^{1},\dots,X^{i-1},X^{i+1},\dots,X^{L})$ and run the maximum entropy-based sampling procedure (Algorithm~\ref{alg:entropy-based-sampling-max}) in the order induced by the permutation
$$\Pi_{\smallsetminus i} \defn \big(\Pi_1, \dots, \Pi_{\Pi^{-1}_i-1}, \Pi_{\Pi^{-1}_i+1}, \dots, \Pi_L \big).$$

This auxiliary sampling process produces a sequence of cumulative unmasking sets $(\cW^{\smallsetminus i}_t)_{t=0}^L$ and a sequence of unmasking sets $(\cD^{\smallsetminus i}_t)_{t=1}^L$.
Here, we slightly abuse notation and continue using $\cW^{\smallsetminus i}$ and $\cD^{\smallsetminus i}$ to denote the cumulative and incremental unmasking sets in the auxiliary process as in Appendix~\ref{sec:proof-entropy-sum-token-i-kl-bound}, but now they are defined based on the maximum entropy-based sampling procedure instead of the entropy sum-based sampling procedure.

As before, we make several key observations.
\begin{enumerate}[label=\arabic*.,ref={\thesection\arabic*}]
    \item \label{item:auxiliary-max-independence} 
For any iteration $t\geq 1$, conditioned on $(\cW^{\smallsetminus i}_{t-1}, X^{\cW^{\smallsetminus i}_{t-1}})$ and $\Pi_{\smallsetminus i}$, the auxiliary sets $(\cD^{\smallsetminus i}_t, \cW^{\smallsetminus i}_t)$ are fully determined. In addition, they are independent of the insertion position $\Pi^{-1}_i$ for token $i$.

\item \label{item:auxiliary-max-subset}
By definition, $\cD^{\smallsetminus i}_{t_i}$ stands for the index set of tokens unmasked at the iteration where token $i$ is unmasked in the auxiliary sampling process.
Because the auxiliary process never considers the token with index $i$, the set $\cD^{\smallsetminus i}_{t_i}$ consists of either all low-entropy indices up to (and including) the first high-entropy index among indices not equal to $i$, or low-entropy indices with size at most $s_{\max}$.
Consequently, the index set of tokens that are unmasked in the same iteration as token $i$ but after it in the original process must be a subset of the auxiliary unmasking set, i.e.,
\begin{align*}
\cL^i = \big\{\Pi_{\Pi^{-1}_i+1}, \dots, \Pi_{|\cW_{t_i}|}\big\} \subseteq \cD^{\smallsetminus i}_{t_i}.
\end{align*}

\item \label{item:auxiliary-max-equal}
By construction, we have $\cW_k = \cW^{\smallsetminus i}_{k}$ and $\cD_k = \cD^{\smallsetminus i}_{k}$ for all $1\leq k < t_i$, and hence 
\begin{align*}
\ptI(X^{i}; X^{\cD^{\smallsetminus i}_{t_i}} \mid \cW_{t_i-1},X^{\cW_{t_i-1}},\Pi_{\smallsetminus i},\Pi_i^{-1}) &= \ptI(X^{i}; X^{\cD^{\smallsetminus i}_{t_i}} \mid \cW^{\smallsetminus i}_{t_i-1},X^{\cW^{\smallsetminus i}_{t_i-1}},\Pi_{\smallsetminus i},\Pi^{-1}_i) 
\\
\ptH(X^{i} \mid \cW_{t_i-1},X^{\cW_{t_i-1}}) &= \ptH(X^{i} \mid \cW^{\smallsetminus i}_{t_i-1},X^{\cW^{\smallsetminus i}_{t_i-1}}). 
\end{align*}
\end{enumerate}

In light of these observations, we can bound the target in \eqref{eq:low-dim-mutual-info-target} as
\begin{align}\label{eq:mutual-info-low-dim-bound}
    &\ptI (X^{i}; X^{\cL^i} \mid \cW_{t_i-1} , X^{\cW_{t_i-1}}, \Pi  )  \ind\big\{\ptH(X^{i} \mid \cW_{t_i-1} , X^{\cW_{t_i-1}}) \leq \eta\big\} \notag \\
    & \quad  \leq 
\ptI(X^{i}; X^{\cD^{\smallsetminus i}_{t_i}} \mid \cW_{t_i-1} , X^{\cW_{t_i-1}}, \Pi  )  \ind\big\{\ptH(X^{i} \mid \cW_{t_i-1} , X^{\cW_{t_i-1}}) \leq \eta\big\}.\notag \\
& \quad = 
\ptI(X^{i}; X^{\cD^{\smallsetminus i}_{t_i}} \mid \cW^{\smallsetminus i}_{t_i-1} , X^{\cW^{\smallsetminus i}_{t_i-1}}, \Pi_{\smallsetminus i}, \Pi^{-1}_i   )  \ind\big\{\ptH\big(X^{i} \mid \cW^{\smallsetminus i}_{t_i-1} , X^{\cW^{\smallsetminus i}_{t_i-1}}) \leq \eta\big\} 
\end{align}
where the second line arises from 
Observation~\ref{item:auxiliary-max-subset} that $\cL^i \subseteq \cD^{\smallsetminus i}_{t_i}$ 
and the monotonicity of mutual information; the last line holds due to Observation~\ref{item:auxiliary-max-equal}.

Therefore, it suffices to control the right-hand side of \eqref{eq:mutual-info-low-dim-bound}. Towards this, we define the random variable
\begin{align}\label{eq:Gk-def}
G_k &\defn \ptI\big(X^{i}; X^{\cD^{\smallsetminus i}_{k}} \mid \cW^{\smallsetminus i}_{k-1}, X^{\cW^{\smallsetminus i}_{k-1}}, \Pi_{\smallsetminus i}, \Pi^{-1}_i  \big)  \ind\big\{\ptH(X^{i} \mid \cW^{\smallsetminus i}_{k-1}, X^{\cW^{\smallsetminus i}_{k-1}}) \leq \eta\big\} \notag\\
&= \ptI\big(X^{i}; X^{\cD^{\smallsetminus i}_{k}} \mid \cW^{\smallsetminus i}_{k-1}, X^{\cW^{\smallsetminus i}_{k-1}}, \Pi_{\smallsetminus i}  \big)  \ind\big\{\ptH(X^{i} \mid \cW^{\smallsetminus i}_{k-1}, X^{\cW^{\smallsetminus i}_{k-1}}) \leq \eta\big\},
\end{align}
for each $k\in[L]$. 
Here, the second line arises from Observation~\ref{item:auxiliary-max-independence}, namely,   conditioned on $\Pi_{\smallsetminus i}$ and $(\cW^{\smallsetminus i}_{k-1},X^{\cW^{\smallsetminus i}_{k-1}})$, $\cD^{\smallsetminus i}_k$ is independent of $\Pi^{-1}_i$.
We can then write the target quantity to be controlled on the right-hand side of \eqref{eq:mutual-info-low-dim-bound} as
\begin{align}
& \ptI(X^{i}; X^{\cD^{\smallsetminus i}_{t_i}} \mid \cW^{\smallsetminus i}_{t_i-1} , X^{\cW^{\smallsetminus i}_{t_i-1}}, \Pi  )  \ind\big\{\ptH(X^{i} \mid \cW^{\smallsetminus i}_{t_i-1} , X^{\cW^{\smallsetminus i}_{t_i-1}}) \leq \eta\big\} = G_{t_i} = \sum_{k=1}^{L}\ind\{t_i = k\} G_{k}, \label{eq:Gti-Gk}
\end{align}



To proceed, writing
\begin{align*}
\bE\big[ G_{t_i}\big] = \bE_{X^{\smallsetminus i}, \Pi_{\smallsetminus i}}\Big[\bE_{\Pi^{-1}_i}\big[ G_{t_i} \mid X^{\smallsetminus i}, \Pi_{\smallsetminus i}\big] \Big], 
\end{align*}
we focus on the inner expectation $\bE_{\Pi^{-1}_i}\big[ G_{t_i} \mid X^{\smallsetminus i}, \Pi_{\smallsetminus i}\big]$, where the expectation is taken over the randomness of $\Pi^{-1}_i$, the insertion position of token $i$ in the permutation $\Pi$, conditioned on $X^{\smallsetminus i}$ and $\Pi_{\smallsetminus i}$.

By Observation~\ref{item:auxiliary-max-independence} and the definition in \eqref{eq:Gk-def}, $G_k$ is deterministic conditioned on $\Pi_{\smallsetminus i}$ and $X^{\smallsetminus i}$ for each $k\in[L]$. Therefore, the inner expectation can be computed as
\begin{align}\label{eq:G-expectation}
\bE_{\Pi^{-1}_i}\big[ G_{t_i} \mid X^{\smallsetminus i}, \Pi_{\smallsetminus i}\big] & = \bE_{\Pi^{-1}_i}\Bigg[ \sum_{k=1}^L G_{k} \ind\{t_i = k\} \mid X^{\smallsetminus i}, \Pi_{\smallsetminus i}\Bigg] = \sum_{k=1}^L G_k \bP \big\{t_i = k \mid X^{\smallsetminus i}, \Pi_{\smallsetminus i} \big\}.
\end{align}


This suggests that we need to study the random iteration index $t_i$ at which token $i$ is unmasked in the original sampling process, which is determined by the insertion position $\Pi^{-1}_i$ of token $i$ in the permutation~$\Pi$. This is accomplished by the following lemma, whose proof is deferred to Appendix~\ref{sec:proof-insertion-position-prob}.
\begin{lemma}\label{lem:insertion-position-prob}
Conditioned on $\Pi_{\smallsetminus i}$ and $X^{\smallsetminus i}$, the probability that token $i$ is unmasked at iteration $k$ in the original sampling process is upper bounded by
\begin{align*}
\mathbb{P}_{\Pi^{-1}_i} \big\{t_i = k \mid X^{\smallsetminus i}, \Pi_{\smallsetminus i} \big\}
\le \frac{s_{\max}}{L},\quad \forall\,k\in[L].
\end{align*}
\end{lemma}

With Lemma~\ref{lem:insertion-position-prob} in hand, we can continue the derivation in \eqref{eq:G-expectation} as
\begin{align*}
\bE_{\Pi^{-1}_i}\big[ G_{t_i} \mid X^{\smallsetminus i}, \Pi_{\smallsetminus i}\big] & = \sum_{k=1}^L G_k \bP \big\{t_i = k \mid X^{\smallsetminus i}, \Pi_{\smallsetminus i} \big\} \\
& \leq \frac{s_{\max}}{L}\sum_{k = 1}^L  G_k \notag\\
& = \frac{s_{\max}}{L}\sum_{k = 1}^L  \ptI(X^{i}; X^{\cD^{\smallsetminus i}_k} \mid \cW^{\smallsetminus i}_{k-1}, X^{\cW^{\smallsetminus i}_{k-1}}, \Pi_{\smallsetminus i} ) \ind\big\{ \ptH(X^{i} \mid \cW^{\smallsetminus i}_{k-1}, X^{\cW^{\smallsetminus i}_{k-1}}) \leq \eta\big\},
\end{align*}
where the last line follows from the definition of $G_k$ in \eqref{eq:Gk-def}. Taking the expectation over $\Pi_{\smallsetminus i}$ and $X^{\smallsetminus i}$ on both sides, we obtain
\begin{align}
\bE[G_{t_i}] 
& \leq \frac{s_{\max}}{L} \bE\Bigg[\sum_{k = 1}^L  \ptI(X^{i}; X^{\cD^{\smallsetminus i}_k} \mid \cW^{\smallsetminus i}_{k-1}, X^{\cW^{\smallsetminus i}_{k-1}}, \Pi_{\smallsetminus i} ) \ind\big\{ \ptH(X^{i} \mid \cW^{\smallsetminus i}_{k-1}, X^{\cW^{\smallsetminus i}_{k-1}}) \leq \eta\big\} \Bigg]. \label{eq:Gti-bound}
\end{align}
Now applying Lemma~\ref{lem:mutual-info-chain-rule-max} with $s = 1$, the summation on the right-hand side of \eqref{eq:Gti-bound} satisfies
\begin{align*}
    \bE\Bigg[\sum_{k = 1}^L  \ptI(X^{i}; X^{\cD^{\smallsetminus i}_k} \mid \cW^{\smallsetminus i}_{k-1}, X^{\cW^{\smallsetminus i}_{k-1}}, \Pi_{\smallsetminus i} ) \ind\big\{ \ptH(X^{i} \mid \cW^{\smallsetminus i}_{k-1}, X^{\cW^{\smallsetminus i}_{k-1}}) \leq \eta\big\} \Bigg]
\leq \ptH (X^i) \wedge \eta \leq \eta.
\end{align*}
Plugging this into \eqref{eq:Gti-bound} and combining it with \eqref{eq:Gti-Gk}, we can bound the expectation of the right-hand side of \eqref{eq:mutual-info-low-dim-bound} by
\begin{align*}
\bE\big[\ptI(X^{i}; X^{\cD^{\smallsetminus i}_{t_i}} \mid \cW^{\smallsetminus i}_{t_i-1} , X^{\cW^{\smallsetminus i}_{t_i-1}}, \Pi  )  \ind\big\{\ptH(X^{i} \mid \cW^{\smallsetminus i}_{t_i-1} , X^{\cW^{\smallsetminus i}_{t_i-1}}) \leq \eta\big\}\big] \leq \frac{s_{\max}}{L} \eta.
\end{align*}

Finally, putting this together with \eqref{eq:mutual-info-low-dim-bound}, we finish the proof of Lemma~\ref{lem:mutual-info-bound-auxiliary-max}.

\subsection{Proof of Lemma~\ref{lem:tau-bound}}
\label{sec:proof-tau-bound}

By the termination condition of the maximum entropy-based sampling procedure, we can split iterations into three types:
\begin{itemize}
    \item cap iterations: end because $D_t=|\cD_t|=s_{\max}$;
	\item high-entropy iterations: end because the stopping token has entropy exceeding $\eta$;
	\item the final iteration, which contributes at most 1 to the total number of iterations $T$.
\end{itemize}
Let $T_{\mathsf{cap}}$ and $T_{\mathsf{high}}$ denote the number of the first two types, respectively. We have $T \leq T_{\mathsf{cap}} + T_{\mathsf{high}} + 1$.

For $T_{\mathsf{cap}}$, since each cap iteration unmasks $s_{\max}$ tokens, we have 
\begin{align}\label{eq:tau-cap-bound}
T_{\mathsf{cap}} \leq L/s_{\max}.
\end{align}
Therefore, the remainder of the proof focuses on controlling $\bE[T_{\mathsf{high}}]$.

Fix an arbitrary $t\in[T]$. Suppose that the $t$-th iteration is a high-entropy iteration. Let $i_t \in \cD_t$ be the index of the last unmasked token at iteration $t$, i.e., $i_t = \Pi_{|\cW_{t}|}$. By the stopping rule, we know that 
\begin{align*}
\eta < \ptH(X^{i_t} \mid \Pi, \cW_{t-1}, X^{\cW_{t-1}}) \leq \ptH(X^{\cD_t} \mid \Pi, \cW_{t-1}, X^{\cW_{t-1}}),
\end{align*}
where the last inequality follows from the monotonicity of entropy. This implies that 
\begin{align*}
\ind\{\text{iteration } t \text{ is high-entropy}\} \leq \frac1\eta \ptH(X^{\cD_t} \mid \Pi, \cW_{t-1}, X^{\cW_{t-1}}).
\end{align*}
Summing over $t\in[T]$ and taking expectation, we obtain
\begin{align}
\bE[T_{\mathsf{high}}] & \leq \frac1\eta \bE\Bigg[\sum_{t=1}^T \ptH(X^{\cD_t} \mid \Pi, \cW_{t-1}, X^{\cW_{t-1}})\Bigg] .
\end{align}

As shown in \eqref{eq:entropy-expression-joint} from Lemma~\ref{lem:entropy-expression}, we know that the sum on the right-hand side of the above display satisfies
\begin{align*}
\bE\Bigg[\sum_{t=1}^T \ptH(X^{\cD_t} \mid \Pi, \cW_{t-1}, X^{\cW_{t-1}})\Bigg] = H(X_0),
\end{align*}
Consequently, we obtain 
\begin{align}\label{eq:tau-high-bound}
\bE[T_{\mathsf{high}}] \leq \frac{H(X_0)}{\eta}
= \frac{L}{s_{\max}},
\end{align}
where the last step is due to our choice $\eta = H(X_0)s_{\max}/L$.

Combining \eqref{eq:tau-cap-bound} and \eqref{eq:tau-high-bound}, we conclude that 
\begin{align*}
\bE[T] \leq \bE[T_{\mathsf{cap}}] + \bE[T_{\mathsf{high}}] + 1 \leq  \frac{2L}{s_{\max}} + 1.
\end{align*}

\subsection{Proof of Lemma~\ref{lem:insertion-position-prob}}
\label{sec:proof-insertion-position-prob}

Fix an arbitrary token index $i\in[L]$ and condition on an arbitrary realization of $\Pi_{\smallsetminus i}$ and $X^{\smallsetminus i}$. 
Recall the auxiliary sampling process defined by removing token $i$.
Conditioned on $\Pi_{\smallsetminus i}$ and $X^{\smallsetminus i}$, the auxiliary unmasking sets $(\cD^{\smallsetminus i}_t, \cW^{\smallsetminus i}_t)_{t\in[L]}$ are deterministic.

Because $\Pi$ is uniformly random over all permutations of $[L]$, after conditioning on the relative order of all indices except $i$, the insertion position $\Pi^{-1}_i$ of token $i$ is uniformly random over $[L]$:
\begin{align*}
\bP \big\{\Pi^{-1}_i =r \mid \Pi_{\smallsetminus i},X^{\smallsetminus i}\big\}=\frac1L,\quad r=1,\dots,L.
\end{align*}

Now fix an arbitrary $k\in[L]$ and consider the event $\{t_i = k\}$, i.e., token $i$ is unmasked at iteration $k$ in the original sampling process. By Observation~\ref{obs:coincide}, the original and auxiliary processes coincide before iteration~$k$. In particular, both processes have unmasked the same set of tokens $\cW^{\smallsetminus i}_{k-1}$. Therefore, token~$i$ must appear after the first $W^{\smallsetminus i}_{k-1} = |\cW^{\smallsetminus i}_{k-1}|$ tokens in the full permutation $\Pi$, implying that the insertion position of token~$i$ satisfies $\Pi^{-1}_i > W^{\smallsetminus i}_{k-1}$. 

Meanwhile, during iteration $k$, before token $i$ is encountered, the original process sees exactly the same previously revealed tokens and the same sequence of candidate indices as the auxiliary process. Hence, before reaching token $i$, it cannot proceed past the stopping point of iteration $k$ in the auxiliary process. In particular, if $i$ were inserted after the block $\cD^{\smallsetminus i}_k$, then the original process would stop iteration $k$ before unmasking token $i$. Thus, token $i$ must be inserted within the block of permutation positions corresponding to $\cD^{\smallsetminus i}_k$, which means that the insertion position of token $i$ satisfies $\Pi^{-1}_i \leq W^{\smallsetminus i}_k = |\cW^{\smallsetminus i}_k|$.

Combining the above two observations, we find 
\begin{align*}
\bP\bigl\{t_i = k \mid \Pi_{\smallsetminus i},X^{\smallsetminus i}\bigr\} & \leq \bP\bigl\{W^{\smallsetminus i}_{k-1} < \Pi^{-1}_i \leq W^{\smallsetminus i}_k \mid \Pi_{\smallsetminus i},X^{\smallsetminus i}\bigr\} = \frac{D^{\smallsetminus i}_k}{L} \leq \frac{s_{\max}}{L},
\end{align*}
where the last step holds because by construction of the maximum entropy-based scheme, each iteration unmasks at most $s_{\max}$ tokens. 

As the above bound holds for any $k\in[L]$, we finish the proof of Lemma~\ref{lem:insertion-position-prob}.

\end{document}